\documentclass{article}

\usepackage{arxiv}

\usepackage[utf8]{inputenc} 
\usepackage[T1]{fontenc}    
\usepackage{hyperref}       
\usepackage{url}            
\usepackage{booktabs}       
\usepackage{amsfonts}       
\usepackage{amsmath}
\usepackage{bm}
\usepackage{nicefrac}       
\usepackage{microtype}      
\usepackage{cleveref}       
\usepackage{lipsum}         
\usepackage{graphicx}
\graphicspath{{./}}
\usepackage{natbib}
\usepackage{doi}
\usepackage{subcaption}
\usepackage{siunitx}
\usepackage{arydshln}		
\newcommand{\NA}{---}
\usepackage{multirow}
\usepackage{bbm}

\title{Fast and reliable uncertainty quantification with neural network ensembles for industrial image classification}

\date{}

\author{ \href{https://orcid.org/0000-0001-9107-5646}{\includegraphics[scale=0.06]{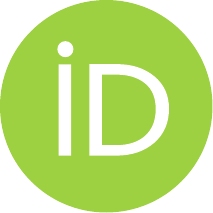}\hspace{1mm}Arthur Thuy}\thanks{Corresponding author} \\
	Ghent University\\
	CVAMO Core Lab, Flanders Make\\
	\texttt{arthur.thuy@ugent.be} \\
	\And
	\href{https://orcid.org/0000-0001-9901-8507}{\includegraphics[scale=0.06]{orcid.pdf}\hspace{1mm}Dries F.~Benoit} \\
	Ghent University\\
	CVAMO Core Lab, Flanders Make\\
	\texttt{dries.benoit@ugent.be} \\
}

\renewcommand{\headeright}{Accepted Manuscript version of an article published in Annals of Operations Research}
\renewcommand{\undertitle}{Accepted Manuscript version of an article published in Annals of Operations Research (\href{https://doi.org/10.1007/s10479-024-06440-4}{\texttt{10.1007/s10479-024-06440-4}})}
\renewcommand{\shorttitle}{}

\hypersetup{
pdftitle={Fast and reliable uncertainty quantification with neural network ensembles for industrial image classification},
pdfsubject={cs.LG},
pdfauthor={Arthur Thuy, Dries F.~Benoit},
pdfkeywords={Neural network ensembles, Computational efficiency, Uncertainty quantification, Out-of-distribution data, Manufacturing},
}

\begin{document}
\maketitle

\begin{abstract}
	Image classification with neural networks (NNs) is widely used in industrial processes, situations where the model likely encounters unknown objects during deployment, i.e., out-of-distribution (OOD) data.
	Worryingly, NNs tend to make confident yet incorrect predictions when confronted with OOD data. 
	To increase the models' reliability, they should quantify the uncertainty in their own predictions, communicating when the output should (not) be trusted.
	Deep ensembles, composed of multiple independent NNs, have been shown to perform strongly but are computationally expensive.
	Recent research has proposed more efficient NN ensembles, namely the snapshot, batch, and multi-input multi-output ensemble.
	This study investigates the predictive and uncertainty performance of efficient NN ensembles in the context of image classification for industrial processes.
	It is the first to provide a comprehensive comparison and it proposes a novel Diversity Quality metric to quantify the ensembles' performance on the in-distribution and OOD sets in one single metric.
	The results highlight the batch ensemble as a cost-effective and competitive alternative to the deep ensemble. 
	It matches the deep ensemble in both uncertainty and accuracy while exhibiting considerable savings in training time, test time, and memory storage.
\end{abstract}




\section{Introduction}\label{sec:intro}

In recent years, neural networks (NNs) have rapidly gained traction in operations research (OR) \citep{mena2023exploiting} and have demonstrated strong performance in handling high-dimensional inputs in image classification for industrial settings \citep{madhav2023weld}.
During deployment, image classifiers inevitably encounter scenarios where the data distribution differs from the training data distribution.
It may specifically find unseen operating conditions, referred to as \textit{shifted} data or covariate shift \citep{quinonero2022dataset}, and unseen objects, referred to as entirely \textit{out-of-distribution (OOD)} data.
For instance, in industrial parts classification, images of known objects in unseen positions or lighting conditions are shifted data; newly developed objects are OOD data.
Furthermore, for example in defect detection, unseen defect types represent OOD data.
However, when confronted with such shifted or OOD data, NNs tend to make confident yet incorrect predictions \citep{guo2017calibration}.
Specifically, an NN classifier may predict an incorrect label despite assigning a high predicted probability to that class.
This behavior is problematic because the model's performance deteriorates over time without warning the practitioner.

To ensure their reliability, NNs should quantify the uncertainty in their own predictions, referred to as model or epistemic uncertainty.	
This notion of epistemic uncertainty is not captured by standard NN classifiers and communicates when an NN’s output should (not) be trusted, crucial for the safe deployment of NNs in the OR domain \citep{thuy2023explainability}.
In this respect, deep ensembles, comprising multiple independent NNs, have emerged as a popular approach to quantify epistemic uncertainty.
Under shifted and OOD data, deep ensembles raise warnings through increased uncertainty, facilitating timely human intervention to prevent misclassifications \citep{ovadia2019can, thuy2023explainability}.

Despite their strong performance, deploying these ensembles in practice is challenging due to the significant computational and memory demands.
First, low inference times are required for a classification system to be relevant in practice. 
For example in a sorting station, the predictions of a parts classifier should be available quickly enough to keep up with the speed of the conveyor belt.
Second, long training times are impractical and allocating substantial resources is hard to justify considering financial constraints and environmental concerns.
Third, limited memory requirements are important when deploying the NN on-device with restricted compute power.
For example, a mobile device with a camera installed over a sorting station should have sufficient compute to perform inference.
Recent fundamental research has proposed more efficient NN ensembles, namely the snapshot ensemble \citep{huang2017snapshot}, batch ensemble \citep{wen2020batchensemble}, and multi-input multi-output (MIMO) ensemble \citep{havasi2020training}.

This study investigates the predictive and uncertainty performance of efficient NN ensembles in the context of image classification for industrial processes.
The experiments are performed on the SIP-17 dataset \citep{zhu2023towards}, a dataset of synthetic images for industrial parts classification.
The contributions are twofold:
(i) it is the first to provide a comprehensive comparison of a single NN, a deep ensemble, and the three efficient NN ensembles; (ii) we propose a Diversity Quality ($DQ_1$) score to quantify the ensembles' performance on the in-distribution (ID) and OOD sets in one single metric, as opposed to two separate metrics in current literature.

Experimental analysis illustrates that batch ensemble offers a robust alternative to deep ensemble, delivering similar performance at a fraction of the computational cost.
In contrast, snapshot and MIMO ensemble lag behind due to either low predictive performance or poor Diversity Quality.

The remainder of the paper is organized as follows.
Section \ref{sec:related_work} gives an overview of related work, Section \ref{sec:uncertainty_ensembles} discusses uncertainty estimation with ensembles and Section \ref{sec:ensembles} presents the efficient ensembling techniques.
In Section \ref{sec:experiments}, the experiments in industrial parts classification are presented and the Diversity Quality score is outlined.
Section \ref{sec:results_discussion} gives the results and discussion. 
Finally, Section \ref{sec:conclusion} provides a conclusion.


\section{Related Work}
\label{sec:related_work}

NN ensembles are commonly used in OR studies to either improve predictive performance or obtain reliable uncertainty estimates.

For predictive performance, NNs are most often used as an ensemble member in heterogeneous ensembles, together with other machine learning models such as Lasso regression, XGBoost, and Support Vector Machines.
Literature ranges over multiple fields, from manufacturing \citep{yang2021deep, wu2022cs} over healthcare \citep{abedin2021deep, baradaran2023ensemble} and finance \citep{du2021forecasting, cui2022carbon, zhang2023carbon, jiang2022two, zhang2022hybrid, li2021entropy, krauss2017deep}, to politics \citep{easaw2023using}.
Furthermore, although less common than heterogeneous ensembles of NNs and machine learning models, some studies use homogeneous ensembles of NNs, i.e. deep ensembles. 
Relevant literature on deep ensembles is in manufacturing \citep{gupta2023deep}, project tendering \citep{bilal2020big}, customer service \citep{zicari2022combining}, healthcare \citep{poloni2022deep}, and tourism \citep{pitakaso2023gamification}.
The studies find that by combining the outputs of several NNs, an ensemble can outperform any of its members.

Despite the extensive use of NN ensembles for predictive performance, uncertainty quantification for NNs is underinvestigated in OR.
Deep ensembles have been used by \citet{thuy2023explainability, han2022out, kim2023unified, prasad2024very, wen2022new} in the context of OOD detection, reducing misclassifications using classification with rejection \citep{mena2021survey}.
In this workflow, uncertain predictions are discarded and the observations are passed on to a human expert for a label.
Note that \cite{wen2022new} employ a type of snapshot ensemble, although its behavior is not compared to other ensemble techniques or a single NN.
Furthermore, \cite{zou2021resilience} use deep ensembles in active learning, guiding labeling efforts based on epistemic uncertainty estimates.
Heterogeneous ensembles have not been used for uncertainty quantification.

Overall, deep ensembles are appealing to practitioners.
Compared to heterogeneous ensembles of NNs and other machine learning models, deep ensembles are more computationally expensive but easier to implement, as only one underlying model needs to be developed.
Furthermore, compared to heterogeneous ensembles of different NN architectures, deep ensembles are computationally cheaper and easier to implement.

Furthermore, classification with rejection based on uncertainty quantification from deep ensembles is preferred over separate rejector-predictor architectures.
That is, such an architecture first uses an anomaly detection algorithm to decide whether to reject the observation; if the observation is not rejected, the predictor is used for inference \citep{hendrickx2024machine}.
A separated rejector is usually simpler to operationalize but often results in sub-optimal rejection performance as there is no information sharing between the rejector and predictor \citep{homenda2014classification}.
Integrated rejection based on NN uncertainty estimates has the additional benefit that the predictor is better, as a deep ensemble typically has stronger predictive performance than a single NN.

\citet{kraus2020deep} report that the lack of epistemic uncertainty quantification in single NNs is a key limiting factor for adoption in the field of OR.
The reliable uncertainty estimates from deep ensembles could increase the relevance of NNs, highlighting the need for more efficient NN ensembles that bring this uncertainty information at a lower computational cost.


\section{Uncertainty Quantification with Ensembles}
\label{sec:uncertainty_ensembles}

\subsection{Aleatoric and epistemic uncertainty}

Uncertainty arises from two sources: aleatoric and epistemic uncertainty \citep{der2009aleatory}.
Aleatoric uncertainty refers to the notion of randomness and is related to the data-measurement process.
This uncertainty is irreducible even if more data is collected.
Epistemic uncertainty accounts for uncertainty in the model parameters and is naturally high for observations outside the training data distribution.
In contrast to aleatoric uncertainty, collecting more data can reduce epistemic uncertainty.
The sum of these two types represents the total uncertainty in a prediction.

NN classifiers capture aleatoric uncertainty in their softmax output distribution, forming a categorical distribution over the class labels.
Capturing epistemic uncertainty requires an ensemble of NN classifiers in order to evaluate the disagreement among the ensemble members.
A naive ensemble of NNs, i.e., a deep ensemble, is an effective but expensive way to obtain reliable epistemic uncertainty estimates \citep{lakshminarayanan2017simple}.

This study focuses on quantifying aleatoric and epistemic uncertainty in convolutional NNs (CNNs) for image classification. 
However, these uncertainties can also be computed in other NN architectures, e.g., Transformer-based NNs for text \citep{thuy2024active}, as well as in other machine learning models using alternative techniques, e.g., Gaussian Processes and Random Forests \citep{hullermeier2021aleatoric}.

\subsection{Deep ensembles}

A deep ensemble \citep{lakshminarayanan2017simple} is an ensemble of $M$ individual NNs, with each member trained independently on the entire dataset using all input features.
The diversity among the members arises from random weight initializations and from randomness in sampling mini-batches of data during training, resulting in different solutions throughout the complex loss landscape.
At test time with test sample $\mathbf{x}^*$, each of the $M$ members performs one forward pass on the input.
The resulting $M$ softmax distributions are averaged to obtain the ensemble prediction $p(\mathbf{y}^* \mid \mathbf{x}^*)$:
\begin{equation}\label{eq:average_ensembles}
	p(\mathbf{y}^* \mid \mathbf{x}^*) \approx \frac{1}{M} \sum_{m=1}^{M} p(\mathbf{y}^* \mid \mathbf{x}^*,\ \bm{\theta}_m).
\end{equation}

Deep ensembles are computationally expensive because their demands increase linearly with the ensemble size $M$.
Computation-wise, each member requires a separate forward pass during training and testing.
Memory-wise, each member requires a copy of their NN weights.
\citet{lakshminarayanan2017simple} experiment with various ensemble sizes, ranging from 5 to 15 members. They find that a small ensemble with 5 members already improves significantly over a single NN, and that using more than 10 members results in limited improvement.

\subsection{Uncertainty decomposition}
\label{subsec:unc_decomposition}

Each NN ensemble, naive or efficient, has $M$ predictions for each test sample $\mathbf{x}^*$.
Based on these predictions we can calculate the total uncertainty, which in turn can be further decomposed in aleatoric and epistemic uncertainty.

First, total uncertainty $\textrm{TU}(\mathbf{x}^*)$ and aleatoric uncertainty $\textrm{AU}(\mathbf{x}^*)$ are approximated using classical information-theoretic measures; then epistemic uncertainty $\textrm{EU}(\mathbf{x}^*)$ is obtained as the difference \citep{depeweg2018decomposition}: 
\begin{align}
	\textrm{TU}(\mathbf{x}^*) &\approx \mathbb{H}\left[\frac{1}{M} \sum_{m=1}^{M} p(\mathbf{y}^* \mid \mathbf{x}^*,\ \bm{\theta}_{m})\right]\\
	\textrm{AU}(\mathbf{x}^*) &\approx \frac{1}{M} \sum_{m=1}^{M} \mathbb{H}\left[p(\mathbf{y}^* \mid \mathbf{x}^*,\ \bm{\theta}_{m})\right]\\
	\textrm{EU}(\mathbf{x}^*) &= \textrm{TU}(\mathbf{x}^*) - \textrm{AU}(\mathbf{x}^*).
\end{align}
Total uncertainty is computed by averaging the predictive distributions (i.e., softmax distributions) over the different members and calculating the entropy $\mathbb{H}$.
Aleatoric uncertainty is computed by calculating the entropy in each member and averaging the entropies.
Intuitively, aleatoric uncertainty measures uncertainty in the softmax output of individual members; epistemic uncertainty measures how much the members deviate.
Naturally, the single NN has zero epistemic uncertainty because it only consists of one member.
For a classification problem with $K$ classes, the maximum total uncertainty in a prediction is $\log(K)$.

Table \ref{tab:decomp} contains three examples in the context of binary classification with $M=4$ ensemble members. 
The middle and bottom rows both have a total uncertainty of \num{1.0} although the predictions are wildly different. 
Therefore, the total uncertainty alone is insufficient to characterize the NN's predictions; decomposition into aleatoric and epistemic uncertainty is necessary. 
\begin{table}[h!]
	\caption{Examples of uncertainty decomposition (adapted from \citet{thuy2023explainability})}\label{tab:decomp}
	\centering
	\begin{tabular}{lcccc}
		\toprule
		Predictions $p(\mathbf{y}^* \mid \mathbf{x}^*,\ \bm{\theta}_m)$ & $p(\mathbf{y}^* \mid \mathbf{x}^*)$ & $\textrm{TU}(\mathbf{x}^*)$ & $\textrm{AU}(\mathbf{x}^*)$ & $\textrm{EU}(\mathbf{x}^*)$\\
		\midrule
		\{(1.\phantom{0}, 0.\phantom{0}), (1.\phantom{0}, 0.\phantom{0}), (1.\phantom{0}, 0.\phantom{0}), (1.\phantom{0}, 0.\phantom{0})\} & (1.\phantom{0}, 0.\phantom{0}) & 0. & 0. & 0.\\
		\{(0.5, 0.5), (0.5, 0.5), (0.5, 0.5), (0.5, 0.5)\} & (0.5, 0.5) & 1. & 1. & 0.\\
		\{(1.\phantom{0}, 0.\phantom{0}), (0.\phantom{0}, 1.\phantom{0}), (1.\phantom{0}, 0.\phantom{0}), (0.\phantom{0}, 1.\phantom{0})\} & (0.5, 0.5) & 1. & 0. & 1.\\
		\bottomrule
	\end{tabular}
\end{table}


\section{Efficient Ensembling Techniques}
\label{sec:ensembles}

This section describes the three efficient ensemble techniques that address the computational burden of deep ensembles.
Table \ref{tab:overview} gives an overview of their improvements over deep ensembles.

\begin{table}[h]
	\caption{Overview of efficient NN ensembles, relative to deep ensemble}\label{tab:overview}
	\centering
	\begin{tabular}{lccc}  
		\toprule%
		Ensemble type & Faster training & Faster evaluation & Less memory \\
		\midrule
		Deep ensemble & & & \\ \hdashline
		Snapshot ensemble  & \checkmark &  &  \\
		Batch ensemble & \checkmark & \checkmark & \checkmark \\
		MIMO ensemble & \checkmark & \checkmark & \checkmark \\
		\bottomrule
	\end{tabular}
\end{table}

\subsection{Snapshot ensemble}

\begin{figure}[t!]
	\centering
	\includegraphics[width=0.5\textwidth]{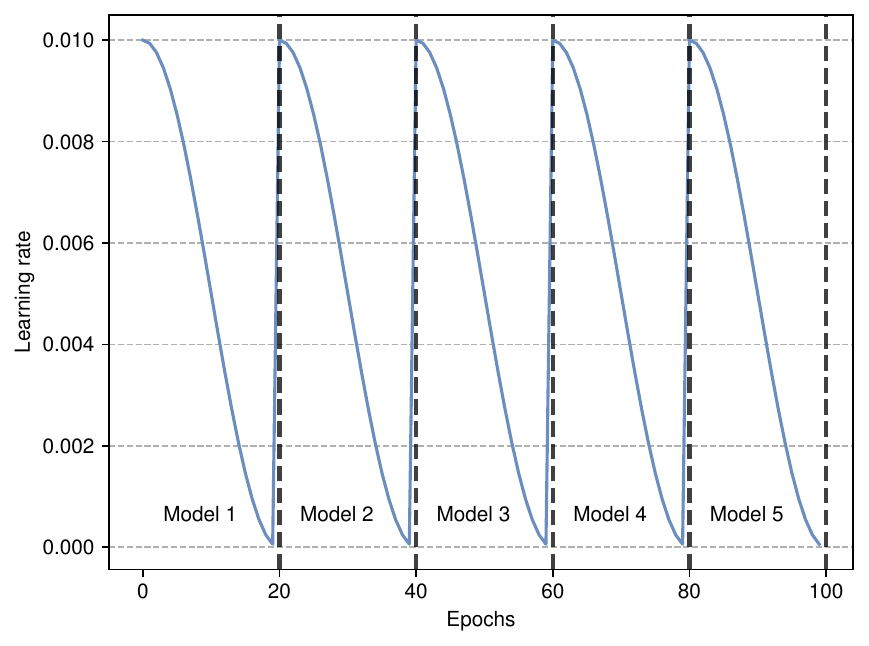} 
	\caption{Snapshot ensemble}
	\label{fig:snapshot_ensemble}
\end{figure}

Snapshot ensemble \citep{huang2017snapshot} trains a single NN and saves the model parameters at different points in time during training.
A cyclic learning rate schedule is used to alternate between converging to local minima and jumping to other modes in the loss landscape (Figure \ref{fig:snapshot_ensemble}).
At each local minimum, the weights are saved and a new ensemble member is acquired.
Model states taken early in the training process have high diversity but poor accuracy, while model states taken later in the training process tend to have high accuracy but are more correlated.
The idea is that the combination of those multiple optima will produce better results than the final model.

The maximum learning rate controls how far the optimizer jumps to another mode after a restart.
Note that the learning rate schedule is determined by the procedure and cannot be combined with another custom learning rate schedule.
The selected number of ensemble members directly affects the training time of each member; too many members damages their individual accuracy while too few members negatively affects the uncertainty estimates.
\citet{huang2017snapshot} find that ensemble sizes of 4--8 work well.

The snapshot ensemble has the same training time as a single NN.
However, similar to deep ensembles, the memory cost and the computational cost at test time increase linearly with the number of ensemble members.

\subsection{Batch ensemble}

\begin{figure}[t!]
	\centering
	\includegraphics[width=0.6\textwidth]{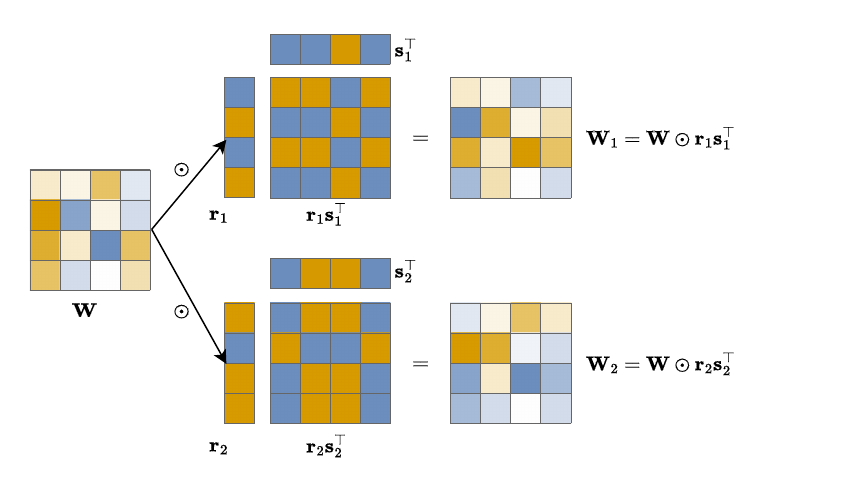} 
	\caption{Batch ensemble (adapted from \citet{wen2020batchensemble})}
	\label{fig:batch_ensemble}
\end{figure}

Instead of storing a separate weight matrix for each ensemble member, batch ensemble \citep{wen2020batchensemble} decomposes the weight matrices into element-wise products of a shared weight matrix and a rank-1 matrix for each ensemble member (Figure \ref{fig:batch_ensemble}).
Let $\mathbf{W} \in \mathbb{R}^{m \times n}$ denote the weights in a NN layer with input dimension $m$ and output dimension $n$.
The ensemble size is $M$ and each ensemble member has weight matrix $\mathbf{W}_i$.
Each member owns trainable vectors $\mathbf{r}_i$ and $\mathbf{s}_i$ which have the same dimension as input $m$ and output $n$, respectively.
Batch ensemble generates the ensemble weights $\mathbf{W}_i$ as follows:
\begin{equation}
	\label{eq:batchensemble}
	\mathbf{W}_i = \mathbf{W} \odot \mathbf{F}_i,\; \text{where}\; \mathbf{F}_i = \mathbf{r}_i \mathbf{s}_i^\top,
\end{equation}
where $\odot$ denotes an element-wise multiplication.
$\mathbf{W}$ is referred to as the \textit{slow} or \textit{shared} weight because it is shared among all ensemble members, and $\mathbf{F}_i$ are the \textit{fast} weights.
During training and testing, the mini-batch is repeated $M$ times which enables all ensemble members to compute the output of the same input data points in one single forward pass.

The fast weights can be initialized with either a random sign vector or a random Gaussian vector.
\citet{wen2020batchensemble} note that the random signed vector often results in higher diversity among the members.
The fast weight can also be updated differently than the slow weights using a learning rate multiplier smaller than 1.
In theory, more ensemble members give better results but a higher computational cost; in this sense, it is similar to the deep ensemble.
\citet{wen2020batchensemble} experiment with ensemble sizes 4 and 8 and find that size 8 generally performs best.

Compared to a single NN, the element-wise product during training and testing is the only additional computation that batch ensemble requires, which is cheap compared to a matrix multiplication.
With respect to memory, the vectors \{$\mathbf{r}_1, \dots, \mathbf{r}_m$\} and \{$\mathbf{s}_1, \dots, \mathbf{s}_m$\} are required, which again is cheap compared to full weight matrices.

\subsection{Multi-input multi-output (MIMO) ensemble}

MIMO ensemble \citep{havasi2020training} builds on the idea of sparsity, as it has been shown that 70--80\% of the NN connections can be pruned without affecting predictive performance \citep{zhu2017prune}.
As such, the MIMO ensemble concurrently trains multiple independent subnetworks within one network, without explicit separation.

\begin{figure}
	\centering
	\begin{subfigure}[b]{0.60\textwidth}
		\centering
		\includegraphics[width=1.0\linewidth]{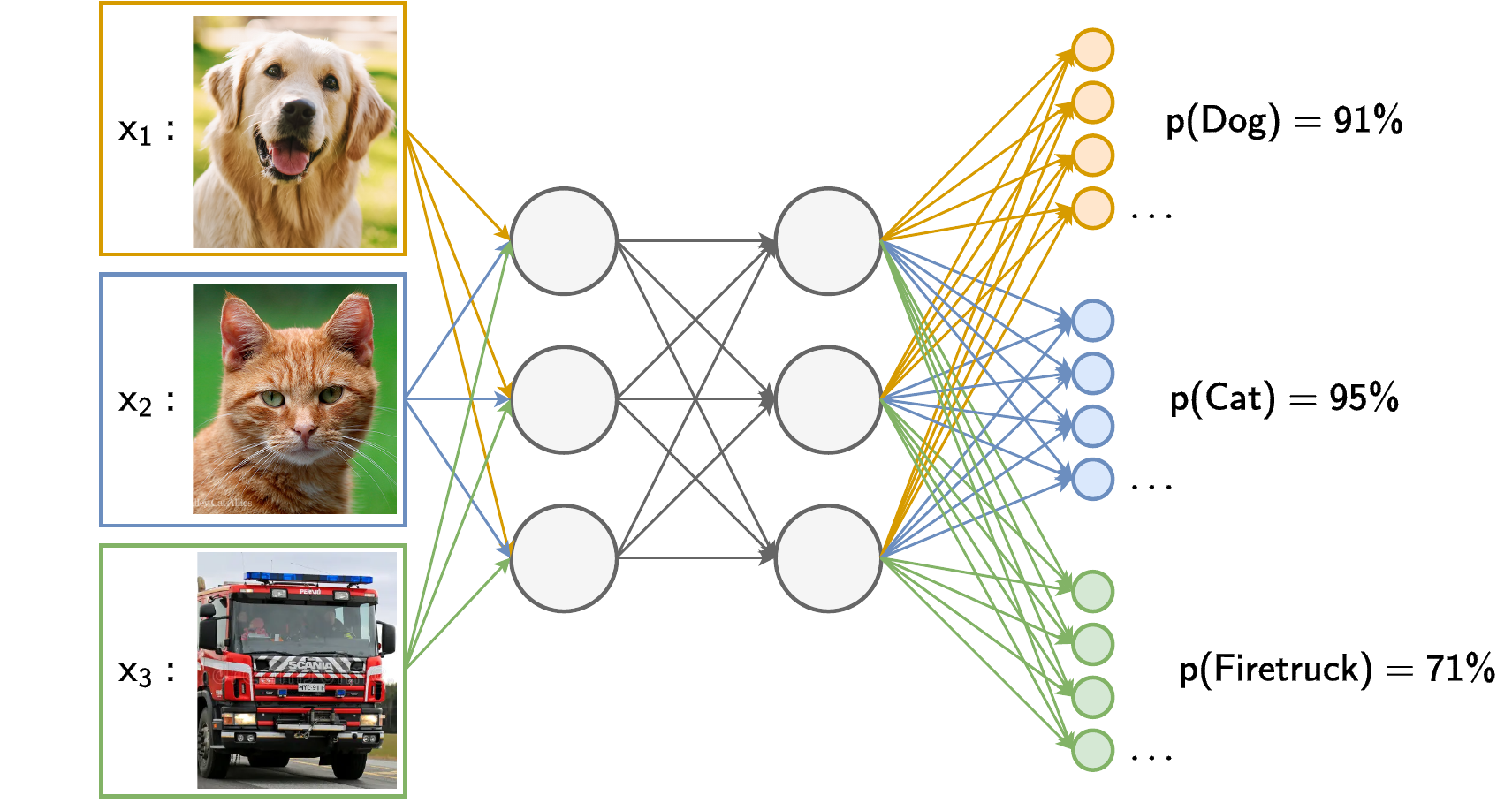}
		\caption{Training}
		\label{fig:mimo_train} 
	\end{subfigure}
	
	\begin{subfigure}[b]{0.75\textwidth} 
		\centering
		\includegraphics[width=1.0\linewidth]{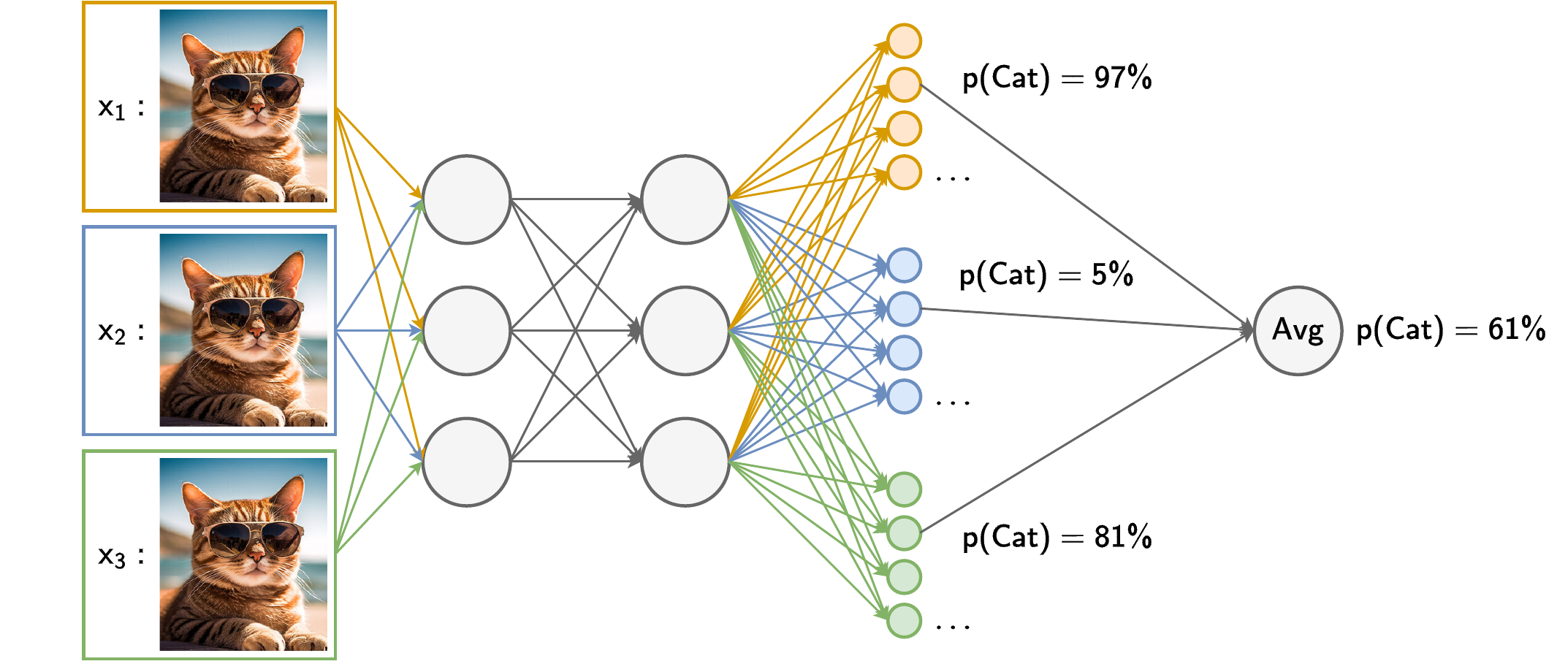}
		\caption{Testing}
		\label{fig:mimo_test}
	\end{subfigure}
	
	\caption{MIMO ensemble at train and test time (adapted from \citet{havasi2020training})}
	\label{fig:mimo_working}
\end{figure}

The MIMO configuration with $M$ members requires two changes to the NN architecture (Figure \ref{fig:mimo_working}).
First, the input layer is replaced, taking $M$ data points instead of a single data point.
Second, the output layer is replaced, consisting of $M$ classification heads based on the last hidden layer instead of a single head.
During training, the $M$ inputs are sampled independently from the training set and each of the $M$ heads is trained to predict its matching input.
At test time, the same input is repeated $M$ times, causing the heads to make $M$ independent predictions on the same input, effectively forming an ensemble.

The number of ensemble members directly affects the number of relationships that need to be learned, which determines the capacity of the subnetworks.
Too many members might make the NN capacity prohibitively small with poor accuracy as a result; too few members will cause poor diversity.
\citet{havasi2020training} report that 3--4 ensemble members is typically optimal.
Additionally, the independence between inputs can be relaxed so that subnetworks can share features.
Relaxing independence might particularly benefit networks with limited excess capacity.
Increasing the input repetition probability improves the predictive performance but negatively impacts the diversity.
Furthermore, examples can be repeated in the minibatch in order to achieve more stable gradient updates, which also has an implicit regularization effect.

The memory and computation cost of the MIMO ensemble is marginally higher than a single NN due to the larger input and output layer.
As such, training, evaluation, and memory costs are slightly higher.


\section{Experiments: Industrial Parts Classification}
\label{sec:experiments}

\subsection{Data}

We use the Synthetic Industrial Parts (SIP-17) dataset \citep{zhu2023towards} containing synthetic images of 17 industrial parts.
Although the images are synthetic, SIP-17 allows for simulating OOD scenarios frequently encountered in real-life OR scenarios.

Large enterprises have inventories reaching thousands of items \citep{ernst1990operations} with 
operational and maintenance support expenses often representing over 60\% of the total costs \citep{hu2015modeling}.
As such, it is infeasible for human operators to manually identify each part and its appropriate inventory strategy.
To this end, identification is often based on barcodes but is prone to wrong labeling, parts may be too small, or functional surfaces do not allow the attachment of a code label.
Using explicit codes is especially challenging in remanufacturing processes of disassembled products, where barcodes are likely dirty or damaged, limiting the automation, reliability, and quality of the processes.
Industrial parts classification offers a solution by identifying parts based on their visual features, maintaining the efficient automation of logistic processes such as part supply and sorted storing.
Depending on the specific setup, a camera can be mounted above a conveyor belt or the camera of a tablet can be used.

The dataset authors organized the images into six separate sub-datasets, each representing a specific use case. 
The first four use cases involve isolated parts requiring classification. 
Use case 1 includes objects mounted in the cabin of a truck, while use cases 2 to 4 feature parts found in various logistics picking stations. 
Accurate classification in these cases ensures that the correct parts are either mounted in the truck cabin or selected for delivery. 
The final two use cases involve assembled objects, where pairs of parts are joined together.
The images were generated by rendering 3D CAD models from varying camera angles and under diverse lighting conditions.

\subsection{Experimental setup}

\begin{figure}[t!]
	\centering
	\includegraphics[width=0.8\textwidth]{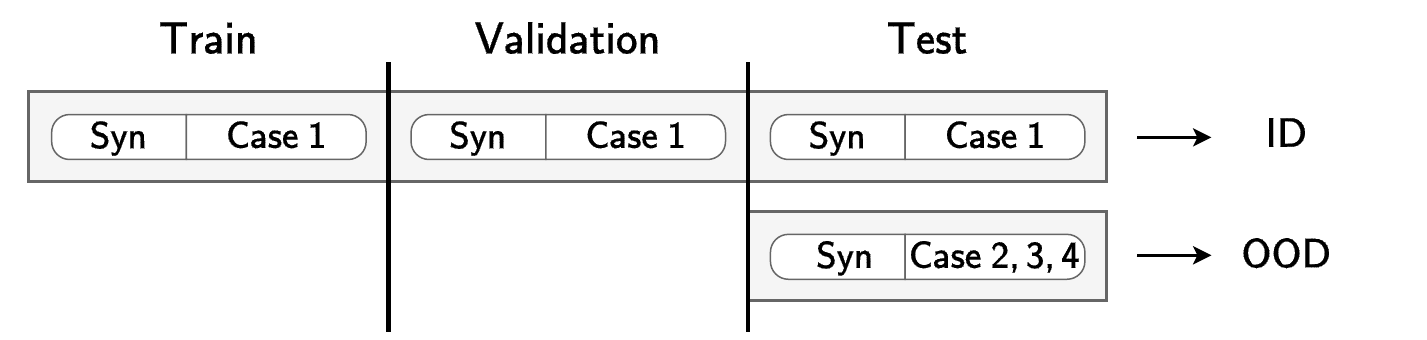} 
	\caption{Experimental setup}
	\label{fig:experimental_setup}
\end{figure}

Figure \ref{fig:experimental_setup} outlines the experimental setup, consisting of an ID and OOD test set.
The models are trained on synthetic images of case 1.
First, we use a synthetic test set of the same distribution, i.e., the ID test set.
Second, we evaluate on synthetic images of the objects in cases 2--4 not observed during training, i.e., the OOD test set.
This set simulates situations such as encountering newly developed parts introduced after the training phase or parts managed by different departments.
It is imperative that the classifier can identify these novel situations and signal a warning, prompting human intervention, rather than making incorrect predictions.

\begin{figure}
	\centering
	\begin{subfigure}[b]{0.60\textwidth}
		\includegraphics[width=1\linewidth]{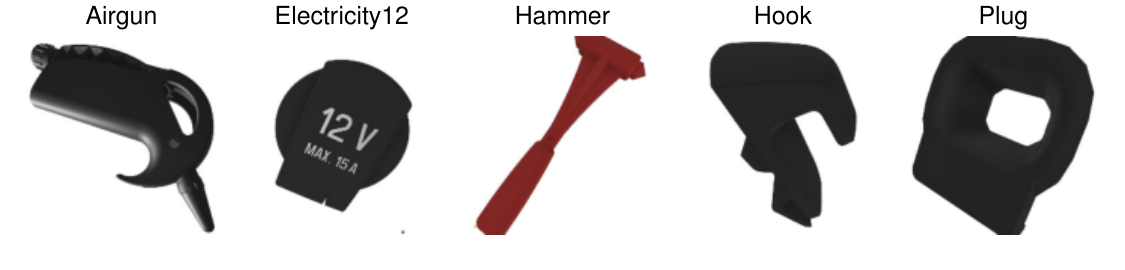}
		\caption{In-distribution set}
		\label{fig:sip_case1} 
	\end{subfigure}
	
	\begin{subfigure}[b]{0.60\textwidth}
		\includegraphics[width=1\linewidth]{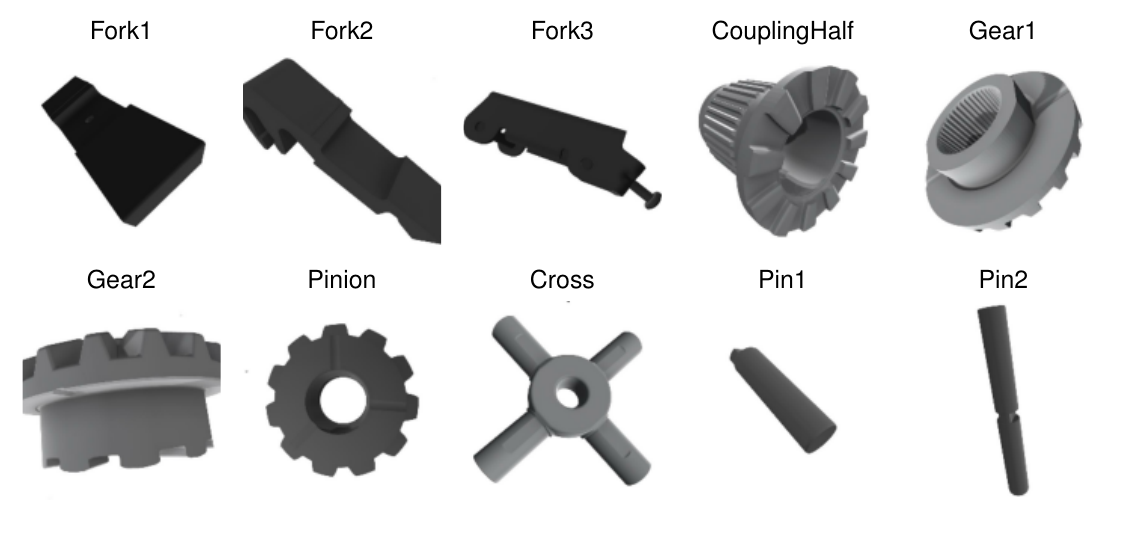}
		\caption{Out-of-distribution set}
		\label{fig:sip_case2}
	\end{subfigure}
	
	\caption{Example observations for the SIP-17 dataset}
	\label{fig:sip_examples}
\end{figure}

\begin{table}[h]
	\caption{Number of images per dataset}\label{tab:dataset_sizes}
	\centering
	\begin{tabular}{lccc}  
		\toprule%
		\textbf{Set} & \textbf{Train} & \textbf{Validation} & \textbf{Test} \\
		\midrule
		ID & 5100 & 900 & 1500\\
		OOD & \NA & \NA & 3000\\
		\bottomrule
	\end{tabular}
\end{table}

Figure \ref{fig:sip_examples} shows an example image for each object type in the two sets.
The ID set consists of 5 industrial parts while the OOD set contains 10 industrial parts; the dataset is balanced.
Note that we do not employ cases 5 and 6 as OOD data because these cases contain images of assembled parts, e.g. the front and rear view of a wheel with and without a wheel nut.
As such, these images are highly similar which would cause the wheel object to be overrepresented in the OOD set.
Table \ref{tab:dataset_sizes} displays the dataset sizes; an overview of the dataset size per object type is available in Appendix \ref{appendix:dataset}.
The ID set available for training is split into a training (85\%) and validation (15\%) set; the OOD set is entirely used as test sets (i.e., 100\%).
We investigate the predictive performance and uncertainty quality of all models, along with their computational expense.

\subsection{Implementation}

Images are resized to 32$\times$32 pixels in grayscale and the dataset is normalized to the [0,1] range.
During training, data augmentation is applied with random horizontal and vertical flips.

Throughout the experiments, the base model is a LeNet-5 with \num{80865} parameters in the standard configuration.
LeNet-5 is a popular CNN architecture which has been shown to work well on smaller input images \citep{chang2023fuzzy}.
Given that for deep and batch ensemble larger ensemble sizes generally translate to better performance, we use $M=8$ members and provide additional results for $M=4$ in Appendix \ref{appendix:all_results}.
For snapshot and MIMO ensemble, selecting the ensemble size $M$ is more complex as using too many or too few members might negatively impact the predictive or uncertainty performance. 
Following the authors' guidelines, we tune snapshot ensembles with 4, 6, and 8 members and select the best-performing configuration based on the validation set. 
Similarly, we tune MIMO ensembles with 3 and 4 members and select the best-performing configuration.
A detailed overview of the hyperparameters is available in Appendix \ref{appendix:hyperparams}, together with the validation performance of each configuration.

The models are implemented using the Uncertainty Baselines repository \citep{nado2021uncertainty} and are trained on an NVIDIA RTX A5000 GPU.
The results are averaged over 5 independent runs with random seeds for a total runtime of 40 hours.

\subsection{Evaluation metrics}

In addition to the classification accuracy $\uparrow$ (the arrow indicates which direction is better), we examine the predictive uncertainty behavior.
Evaluating the predictive uncertainties is challenging as the ``ground truth'' uncertainty estimates are unavailable.
In this work, we calculate the calibration metric negative log-likelihood (NLL) on the ID set.
However, this cannot be calculated on the OOD set because there is no ground truth label and the predictions are incorrect by definition.
Uncertainty quantification can be evaluated using classification with rejection and the novel Diversity Quality metric.

\textbf{Negative log-likelihood (NLL)} $\downarrow$ NLL is a proper scoring rule and a popular metric for evaluating predictive uncertainty \citep{ovadia2019can}.
For $N$ observations and $K$ output classes, this gives:
\begin{equation}
	\textrm{NLL} = -\frac{1}{N} \sum_{n=1}^{N} \sum_{k=1}^{K} \mathbbm{1}(y_n = k) \log p(y = k \mid \mathbf{x}_n,\ \bm{\theta}).
\end{equation}
A lower NLL score indicates a closer alignment of predicted probabilities with the true labels.

\textbf{Non-rejected accuracy (NRA) curves} $\nwarrow$ Classification with rejection assists the practitioner in deciding which high uncertainty predictions should be rejected and passed on to a human expert for labeling.
Accordingly, the accuracy is computed only for observations with a total uncertainty value (Section \ref{subsec:unc_decomposition}) below a user-defined threshold, referred to as non-rejected accuracy (NRA).
We evaluate the NRA score for the entire range of uncertainty values, resulting in an NRA curve. 
Based on the NRA curve for the test sets, the practitioner can decide which absolute uncertainty threshold to use during deployment.

\textbf{Diversity Quality metric} $\uparrow$ To gain a deeper understanding of the ensembling techniques, we assess the diversity among ensemble members.
The diversity metric measures the fraction of test data points on which predictions of ensemble members disagree \citep{fort2019deep}.
This metric is 0 when two members make identical predictions and 1 when predictions differ on every example in the test set.
As the base model for diversity computation, we average the output distributions over the members and determine the resulting predicted label.
The diversity score is related to the epistemic uncertainty in Section \ref{subsec:unc_density} and allows assigning a score to each member.

We propose a novel Diversity Quality ($DQ_1$) score to represent the diversity performance on both the ID and OOD set in one single metric, made available in the Python package \texttt{reject} \citep{Thuy_Reject_2024}.
The ID diversity (IDD) should be as close as possible to 0.0 and the OOD diversity (OODD) as close as possible to 1.0.
The $DQ_1$-score is the harmonic mean of (1 - IDD) and OODD, displayed in Equation \ref{eq:dv_1}. 
The harmonic mean assigns equal weights to both IDD and OODD and is most appropriate to handle ratios \citep{agrrawal2010using}.
It has a minimum value of 0.0 and a maximum value of 1.0.
The new evaluation score has the advantage of being close to 0.0 when either the IDD or OODD is poor, while at the same time, the score is close to 1.0 only when both the IDD and OODD are strong.
\begin{align}\label{eq:dv_1}
	DQ_1 &= 2 \cdot \frac{(1 - IDD) \cdot OODD}{(1 - IDD) + OODD}
\end{align}

Being a harmonic mean, it is similar in construction to the $F_1$-score between precision and recall often used in machine learning literature \citep{sokolova2009systematic}.
The $DQ_1$-score can also be generalized to a $DQ_\beta$-score, valuing one of ID diversity or OOD diversity more than the other.
With $\beta$ a positive real factor, OODD is considered $\beta$ times as important as IDD:
\begin{align}\label{eq:dv_beta}
	DQ_\beta &= (1 + \beta^2) \cdot \frac{(1 - IDD) \cdot OODD}{\beta^2 \cdot (1 - IDD) + OODD}.
\end{align}
For $\beta=1$, Equation \ref{eq:dv_beta} evaluates to Equation \ref{eq:dv_1}. For example, for $\beta=2$, OODD is twice as important as IDD; for $\beta=0.5$, IDD is twice as important as OODD.

The $DQ_1$-score is a valuable tool for practitioners. 
It offers a comprehensive evaluation of model diversity by integrating two closely related metrics.
While practitioners can assess the IDD and OODD metrics individually, this process is mentally demanding and prone to error.

Moreover, the $DQ_{\beta}$-score is flexible as the $\beta$ parameter can be adjusted based on the specific application, enabling prioritization of one metric over the other.
Management can determine an appropriate $\beta$ value by trading off the cost of an incorrect OOD prediction and the cost of consulting a human expert.
For instance, in medical applications where incorrect predictions have severe consequences, practitioners can increase the $\beta$ value to emphasize strong OODD performance.
Effectively flagging OOD observations with high diversity is essential in such cases, as unseen medical situations demand human intervention for reliable diagnoses.

Additionally, when practitioners assess IDD and OODD metrics separately to guide model selection, they implicitly assume a weighting between these metrics. The $DQ_{\beta}$-score formalizes this weighting through the explicit $\beta$ parameter, resulting in a more transparent and reproducible decision-making process.


\section{Results and Discussion}
\label{sec:results_discussion}

This section examines the results for both the ID and OOD sets, focusing on predictive performance, uncertainty behavior, Diversity Quality, and computational cost. 
Finally, we address the trade-offs among these aspects to identify the most suitable NN ensemble for practical applications.
We work with a snapshot ensemble of $M=8$ members and a MIMO ensemble of $M=3$ members because this yields the lowest validation NLL for their ensemble type.
Please refer to Appendix \ref{appendix:hyperparams} for an overview of the validation performance.
Furthermore, we set $\beta=1$ in the $DQ_{\beta}$-score and provide an analysis for a varying $\beta$ in Appendix \ref{appendix:beta_sensitivity}.

\subsection{In-distribution performance}
\label{subsec:id_performance}

We evaluate the classification accuracy and NLL for the ID predictions.
We expect the ensemble methods to perform better than the single NN.
Table \ref{tab:id_predictive} shows the results.

\begin{table}[h]
	\caption{Performance on the ID set}\label{tab:id_predictive}
	\centering
	\begin{tabular}{lcc}  
		\toprule%
		\textbf{Model} & \textbf{Accuracy (\%)} $\uparrow$ & \textbf{NLL} $\downarrow$ \\
		\midrule
		Single & 97.72 $\pm$ 0.14 & 0.1047 $\pm$ 0.0089 \\
		Deep E ($M=8$) & 98.53 $\pm$ 0.09 & 0.0440 $\pm$ 0.0011 \\
		Snapshot E ($M=8$) & 97.15 $\pm$ 0.10 & 0.1048 $\pm$ 0.0031 \\
		Batch E ($M=8$) & 98.65 $\pm$ 0.05 & 0.0504 $\pm$ 0.0035 \\
		MIMO E ($M=3$) & 96.63 $\pm$ 0.30 & 0.1057 $\pm$ 0.0061 \\
		\bottomrule
	\end{tabular}
\end{table}

In terms of accuracy, the deep ensemble outperforms the single neural network (NN) significantly. 
Among the efficient ensembles, both the snapshot ensemble and MIMO ensemble underperform, delivering worse results compared to the single NN. 
In contrast, the batch ensemble demonstrates strong performance, matching that of the deep ensemble. 
Similar trends are observed for the NLL metric, where the batch ensemble again matches the deep ensemble, while the snapshot and MIMO ensembles perform similarly to the single NN. 
Overall, the batch ensemble shows highly promising results on the ID set, equaling the benchmark set by the deep ensemble. 
The following sections will explore the models' behavior on OOD data.

\subsection{Uncertainty density}
\label{subsec:unc_density}

\begin{figure}
	\centering
	\begin{subfigure}[b]{1.0\textwidth}
		\includegraphics[width=1\linewidth]{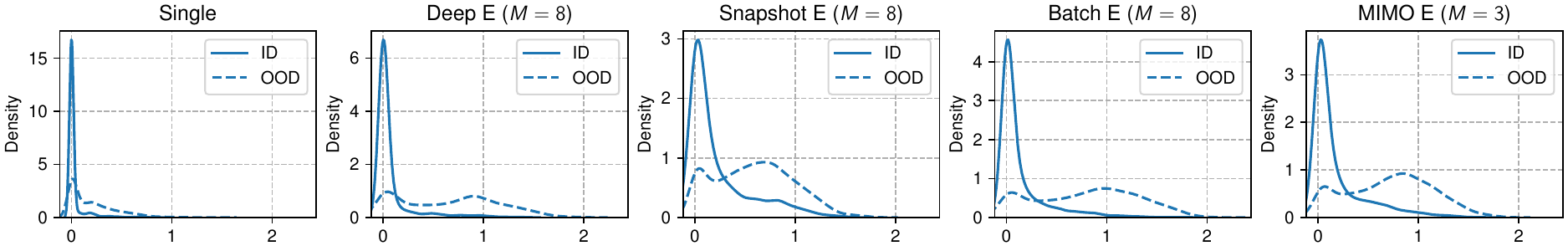}
		\caption{Total uncertainty}
		\label{fig:unc_density_tu} 
	\end{subfigure}
	
	\begin{subfigure}[b]{1.0\textwidth}
		\includegraphics[width=1\linewidth]{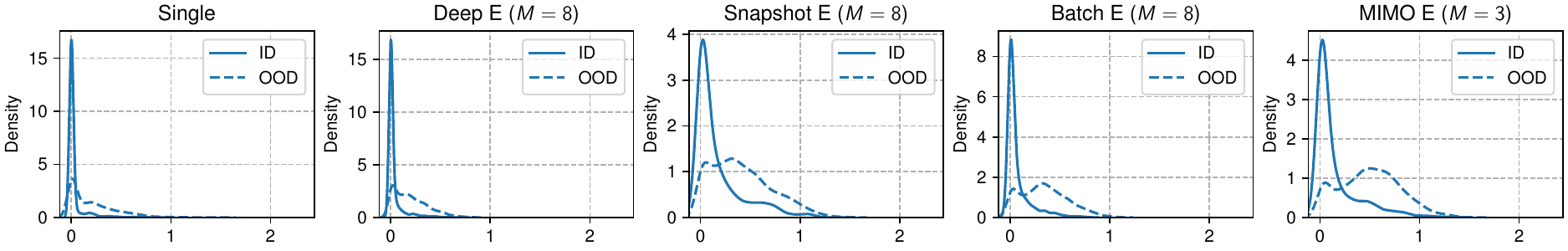}
		\caption{Aleatoric uncertainty}
		\label{fig:unc_density_au}
	\end{subfigure}
	
	\begin{subfigure}[b]{1.0\textwidth}
		\includegraphics[width=1\linewidth]{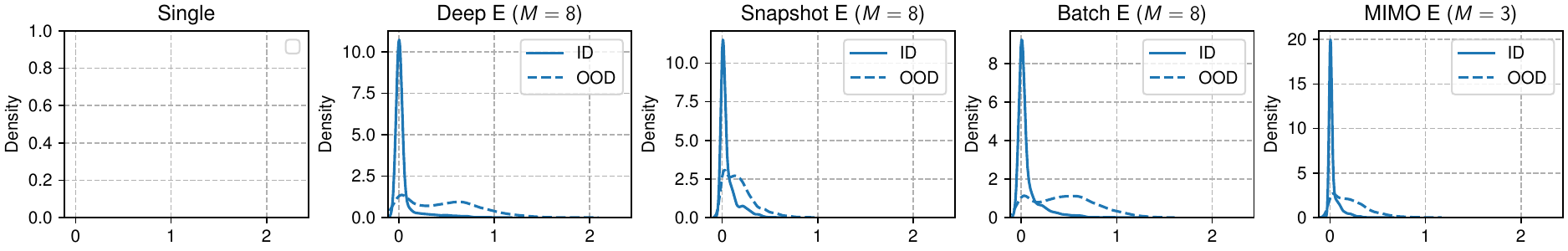}
		\caption{Epistemic uncertainty}
		\label{fig:unc_density_eu}
	\end{subfigure}
	
	\caption{Densities for the aleatoric, epistemic, and total uncertainty on the ID and OOD set}
	\label{fig:unc_density}
\end{figure}

Figure \ref{fig:unc_density} illustrates the density of total, aleatoric, and epistemic uncertainty.
The results are obtained by evaluating the models on the ID and OOD sets, and calculating the uncertainty in each prediction.
Note that total uncertainty is the sum of aleatoric and epistemic uncertainty and has a maximum value of $\log_2(5) = 2.32$.
We do not use the NLL here because this requires ground truth labels unavailable for the OOD set.
On the ID data, we would like to see low total uncertainty.
Conversely, on OOD data, we expect the models to exhibit higher epistemic uncertainty and thus total uncertainty, reflecting an awareness that they know what they do not know.
In a highly idealized scenario, the mode of the total uncertainty lies around zero for the ID set and approaches the maximum value of 2.32 for the OOD set.

The single NN displays low uncertainty on the ID set but only slightly higher uncertainty on the OOD set.
Notably, this uncertainty stems solely from aleatoric uncertainty with a mode around 0.0, as the single NN lacks the ability to capture epistemic uncertainty.
In contrast, the deep ensemble aligns with expectations, demonstrating low uncertainty on the ID set and substantial epistemic uncertainty on the OOD set, resulting in higher total uncertainty with a mode around value 0.95.

Among the efficient ensembles, we observe variations in uncertainty behavior.
The snapshot ensemble exhibits moderately high total uncertainty on the ID set.
Furthermore, total uncertainty increases on the OOD set with a mode around 0.75 but this primarily stems from aleatoric uncertainty, not epistemic uncertainty which is undesirable.
The batch ensemble performs well, mirroring the behavior of the deep ensemble with low uncertainty on the ID set and increased epistemic uncertainty on the OOD set resulting in a mode of total uncertainty around 1.0.
Lastly, the MIMO ensemble exhibits decent uncertainty values on both the ID and OOD sets.
Here, the total uncertainty on the OOD set has a mode around value 0.85 but this is mainly driven by higher aleatoric uncertainty while epistemic uncertainty only experiences a slight increase.

Overall, the batch ensemble once again stands out as the most promising, exhibiting similar uncertainty behavior to the deep ensemble.
Conversely, the snapshot and MIMO ensemble do not show substantially increased epistemic uncertainty on the OOD set.

\subsection{Classification with rejection}
\label{subsec:class_rejection}

\begin{figure}[t!]
	\centering
	\includegraphics[width=0.6\textwidth]{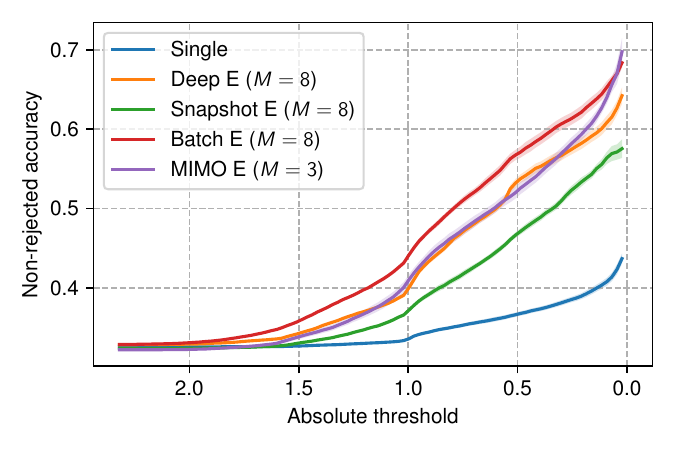} 
	\caption{Non-rejected accuracy}
	\label{fig:nra}
\end{figure}

We evaluate the models on the combination of the ID and OOD test sets.
A model proficient in identifying OOD data exhibits an NRA curve closer to the top-left corner.
Essentially, as the NRA curve rises early and monotonically, the model demonstrates higher uncertainty on observations that in fact prove to be OOD data.
Consequently, it prioritizes rejecting incorrectly classified ID observations or OOD observations before rejecting the correctly classified ID observations.
Figure \ref{fig:nra} displays the NRA for all models based on their total uncertainty.

The single NN exhibits the weakest performance, characterized by a slow increase in NRA, reaching only 45\% at low uncertainty thresholds.
This finding aligns with the low uncertainty observed on the OOD set in Figure \ref{fig:unc_density}.
In contrast, the deep ensemble demonstrates significantly better performance, showing a much earlier and sharper increase in NRA, reaching 65\% at low uncertainty thresholds.

The snapshot ensemble consistently underperforms the deep ensemble.
The batch ensemble demonstrates strong performance, significantly outperforming the deep ensemble across the entire range of thresholds.
The MIMO ensemble closely matches the deep ensemble and outperforms it for smaller uncertainty thresholds.

In summary, the single NN produces confident incorrect predictions, as indicated by the low NRA curve.
The deep ensemble is substantially better at identifying and rejecting OOD observations, demonstrated by its early and steep increase in NRA.
Both batch and MIMO ensemble perform well and match or even improve on the deep ensemble.
The snapshot ensemble falls short of achieving equally strong NRA values.

\subsection{Diversity analysis}
\label{subsec:diversity}

\begin{figure}[t!]
	\centering
	\begin{subfigure}{.5\linewidth}
		\centering
		\includegraphics[width=1.0\textwidth]{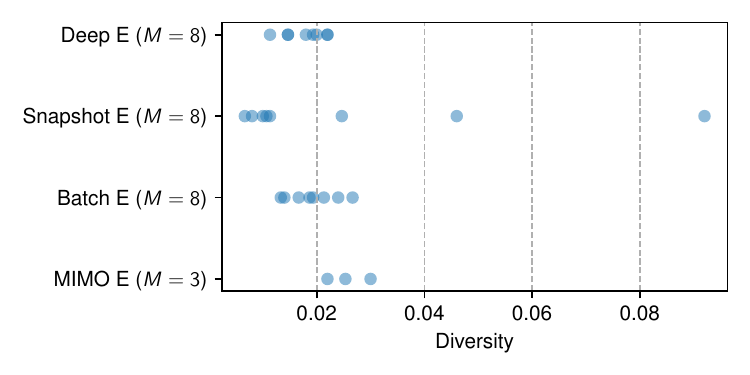} 
		\caption{ID set}
		\label{fig:diversity_id}
	\end{subfigure}%
	\begin{subfigure}{.5\linewidth}
		\centering
		\includegraphics[width=1.0\textwidth]{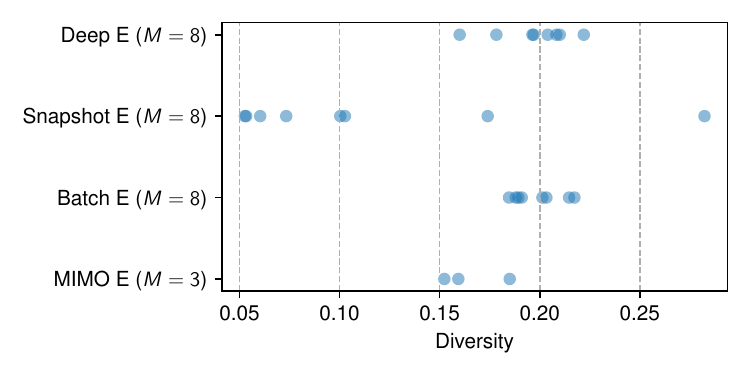} 
		\caption{OOD set}
		\label{fig:diversity_ood}
	\end{subfigure}
	\caption{Diversity on the ID and OOD set.
		Each point represents the diversity of a trained model against the model averaged over all members}
	\label{fig:diversity_id_ood}
\end{figure}

Figure \ref{fig:diversity_id_ood} displays a scatterplot of diversity scores on the ID and OOD test set, where each point represents an ensemble member.
Recall that we want low diversity on the ID set and high diversity on the OOD set.

On the ID set (Figure  \ref{fig:diversity_id}), the deep and batch ensembles perform best and have the lowest diversity, closely followed by the MIMO ensemble.
The snapshot ensemble exhibits a skewed diversity distribution, with most members having very low diversity but others having very high diversity. 
This behavior stems from the initial snapshots being highly diverse and gradually converging as training progresses.

Moving to the OOD set (Figure \ref{fig:diversity_ood}), the diversity scores are indeed orders of magnitude higher.
Here, as on the ID set, the snapshot ensemble still has low diversity for most of its members and high diversity for a few early snapshots, which is undesirable as the diversity should be as high as possible on the OOD set.
The deep and batch ensemble have the highest diversity and perform strongest, closely followed by the MIMO ensemble.

Examining the ID and OOD scenarios independently poses difficulties in obtaining a comprehensive view of diversity performance. 
The proposed metric, Diversity Quality ($DQ_1$), addresses this issue and enables a clear assessment of diversity performance across various ensembles.

\begin{figure}[t!]
	\centering
	\includegraphics[width=.5\textwidth]{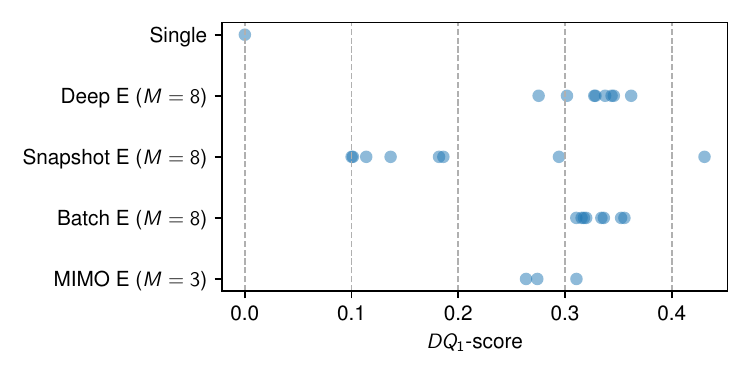} 
	\caption{$DQ_1$-scores. Each point represents the $DQ_1$-score of an ensemble member}
	\label{fig:dq_1_score}
\end{figure}

Figure \ref{fig:dq_1_score} shows the $DQ_1$-scores for each ensemble and its members.
The deep and batch ensemble stand out with the highest $DQ_1$-scores, attributed to their relatively low IDD and high OODD.
Although both ensembles exhibit similar behavior, the deep ensemble has more variation in the diversity scores.
The snapshot ensemble ranks lowest in $DQ_1$-scores because of its undesirable low OODD, despite having low IDD for most members.
This lack of diversity is reflected in its overall poor $DQ_1$-scores. 
Furthermore, the MIMO ensemble slightly underperforms deep and batch ensemble because its IID and OODD are slightly worse.
Note that the single NN has a minimum score of zero because it only has one member, thereby lacking diversity altogether.
Overall, the $DQ_1$-scores comprehensively evaluate the ensembles’ diversity performance.

\subsection{Computational cost}
\label{subsec:compute_cost}

\begin{figure}[t!]
	\centering
	\begin{subfigure}{.5\linewidth}
		\centering
		\includegraphics[width=1.0\textwidth]{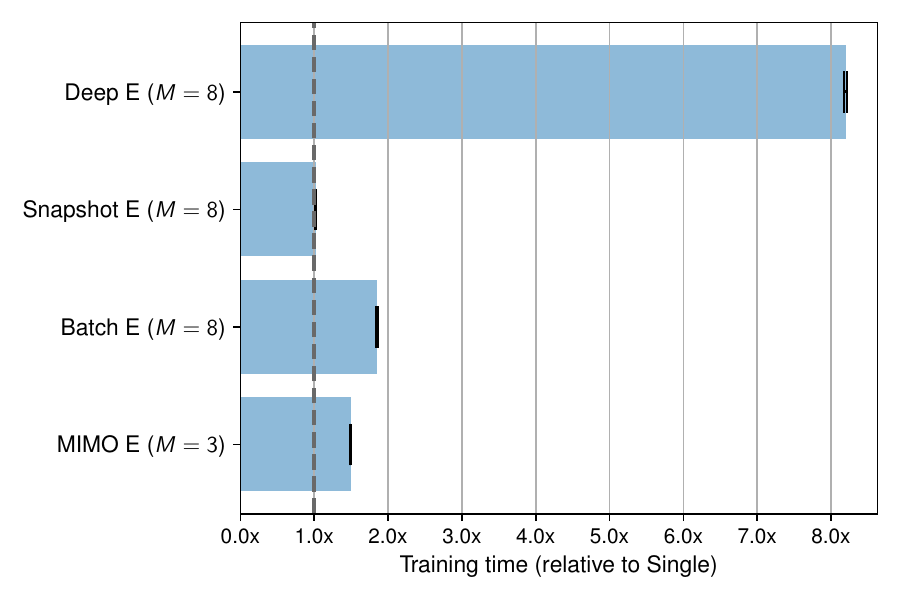} 
		\caption{Training cost}
		\label{fig:train_cost}
	\end{subfigure}%
	\begin{subfigure}{.5\linewidth}
		\centering
		\includegraphics[width=1.0\textwidth]{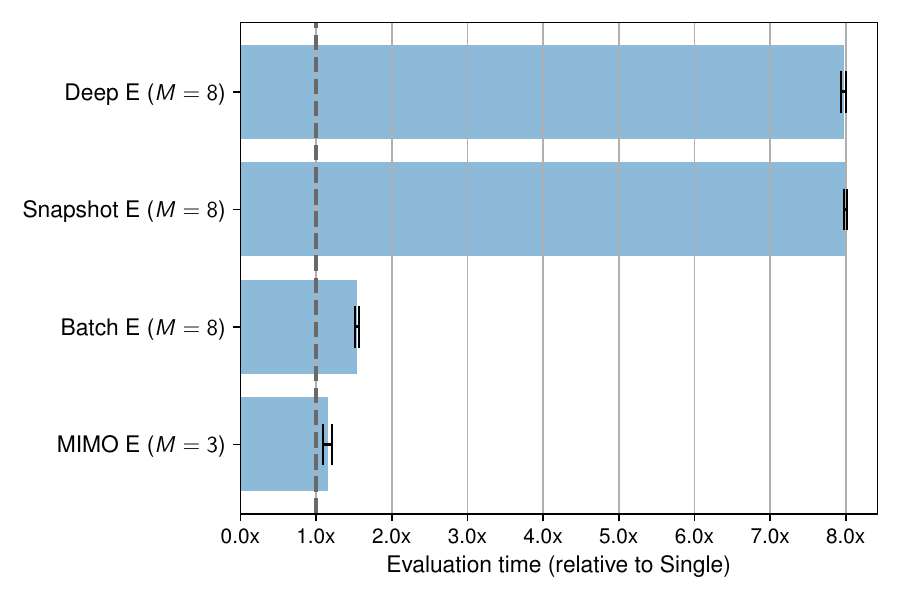} 
		\caption{Evaluation cost}
		\label{fig:eval_cost}
	\end{subfigure}\\[1ex]
	\begin{subfigure}{.5\linewidth}
		\centering
		\includegraphics[width=1.0\textwidth]{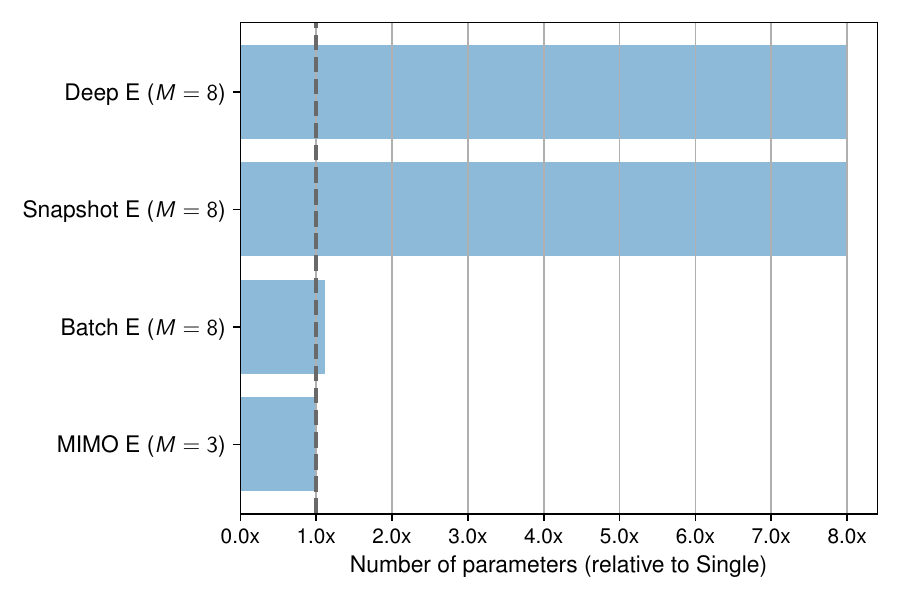} 
		\caption{Memory cost}
		\label{fig:memory_cost}
	\end{subfigure}
	
	\caption{Training, evaluation, and memory cost}
	\label{fig:compute_all}
\end{figure}

Figure \ref{fig:compute_all} depicts the computational costs associated with all ensembling techniques.
The values are normalized relative to the single NN to facilitate easier comparisons.

In Figure \ref{fig:train_cost}, the training cost is presented.
The training time for the deep ensemble with $M=8$ members is 8 times longer than that of the single NN, as it involves training 8 NNs independently.
The training time for the snapshot ensemble is equivalent to that of a single NN, and the number of ensemble members does not impact the training duration.
Both the batch ensemble and MIMO ensemble exhibit slightly slower training times compared to the single NN, with a factor of 1.85x and 1.50x, respectively.
It is important to note that the batch ensemble and MIMO ensemble are trained for 50\% more epochs due to slower learning convergence.

In Figure \ref{fig:eval_cost}, evaluation times are presented.
The deep and snapshot ensembles incur 8 times higher evaluation costs than the single NN because they require $M=8$ independent forward passes.
Conversely, the batch ensemble and MIMO ensemble are only around 50\% and 15\% slower in evaluation, respectively.

Figure \ref{fig:memory_cost} illustrates the number of parameters stored.
Memory consumption for both the deep and snapshot ensembles is 8 times higher, given the weights storage for $M=8$ individual NNs.
The batch ensemble has approximately 10\% more parameters due to the inclusion of additional fast weights.
In the MIMO model, the extra parameters in the input and output layers are negligible.

In summary, the batch ensemble and MIMO ensemble incur only slight additional costs compared to the single NN, making them orders of magnitude more cost-effective than the deep ensemble.

\subsection{Cost-performance analysis}
\label{subsec:cost_performance}

\begin{figure}[t!]
	\centering
	\includegraphics[width=0.6\textwidth]{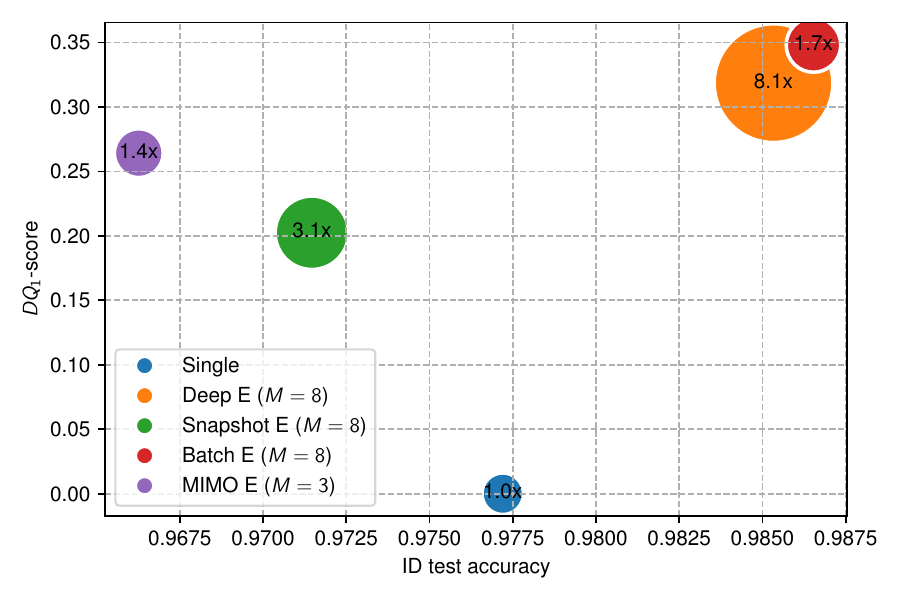} 
	\caption{Accuracy on the ID set and $DQ_1$-score.
		Each point represents an ensemble model; the point size represents the weighted computational cost relative to a single NN.
		Models in the upper-right corner perform best}
	\label{fig:cost_performance}
\end{figure}

Figure \ref{fig:cost_performance} displays a bubble chart combining the predictive accuracy, $DQ_1$-scores, and computational costs of all models described in previous sections.
The horizontal axis shows the test accuracy on the ID set (Section \ref{subsec:id_performance}), where higher scores are better.
The vertical axis shows the $DQ_1$-scores (Section \ref{subsec:diversity}), where higher scores are again better.
The point size shows the weighted computational cost of the models relative to the single NN (Section \ref{subsec:compute_cost}), with smaller points being better.
Hence, optimal models are denoted by small points positioned in the upper-right corner.

To facilitate comparisons, the computational costs of each model are summarized by computing a weighted average of the three individual criteria.
The criteria weights can be selected depending on the specific application at hand.
For industrial parts classification, we allocate 70\% weight to training time, 20\% weight to evaluation time, and 10\% weight to parameter count.
Alternatively, in settings characterized by rapidly moving production lines, evaluation time might receive a higher weight to facilitate timely human intervention.

The plot illustrates that deep and batch ensemble significantly outperform the other models, boasting high accuracy on the ID set and high Diversity Quality.
Batch ensemble, however, is much more computationally efficient than deep ensemble.
It achieves strong performance with few resources by sharing information between the ensemble members, unlike a deep ensemble consisting of entirely independent members.
Furthermore, it is relatively easy to optimize as more ensemble members generally translate to better performance, unlike snapshot and MIMO ensembles.
As such, batch ensemble is a cost-effective and competitive alternative to deep ensemble.
Conversely, the MIMO ensemble demonstrates decent Diversity Quality but has low ID accuracy, failing to match even the single NN.
The snapshot ensemble lags behind the other ensembles in both aspects.


\section{Conclusion}
\label{sec:conclusion}

NN image classifiers deployed in industrial settings are likely to be confronted with OOD scenarios, where they often make confident yet incorrect predictions.
Deep ensembles have been shown to overcome these challenges by quantifying uncertainty in their own predictions but are computationally expensive.
This study therefore investigates more efficient NN ensembles in order to obtain uncertainty estimates at a lower cost, applied on a dataset of industrial parts classification.
It is the first to provide a comprehensive comparison among the methods and it proposes a novel Diversity Quality score to quantify the ensembles' performance on the ID and OOD sets in one single metric.

The superior performance of deep and batch ensemble over the single NN highlights the need for reliable uncertainty estimation to safely deploy NNs in operational settings. 
By sharing information between the ensemble members, batch ensemble emerges as a cost-effective and competitive alternative to deep ensembles.
It matches the deep ensemble both in terms of accuracy and Diversity Quality while demonstrating savings in training time, test time, and memory storage.
Additionally, it exhibits similar uncertainty behavior when confronted with OOD data, and selecting the ensemble size is equally straightforward: more members generally results in better performance.

Several directions for future work are possible.
Expanding studies on industrial parts classification to datasets containing more object classes would validate the findings and better simulate real-world industrial scenarios.
Furthermore, ensembles hold potential in various other OR domains. 
For instance, in tasks like household waste sorting or quality control within production plants, the emergence of new product types or failure modes requires reliable uncertainty in classifiers.
Finally, it is promising to investigate sim-to-real setups \citep{tobin2017domain} where the NN is only trained on synthetic images and evaluated on real images, requiring less collecting and labeling of real images.
As such, this workflow exhibits an inherent distribution shift, also requiring reliable uncertainty quantification.

\section*{Acknowledgments}

The authors would like to thank Thomas Suys for his contributions during his master's thesis. 
This study was supported by the Research Foundation Flanders (FWO) (grant number 1S97022N).




\begin{thebibliography}{52}
	\providecommand{\natexlab}[1]{#1}
	\providecommand{\url}[1]{\texttt{#1}}
	\expandafter\ifx\csname urlstyle\endcsname\relax
	\providecommand{\doi}[1]{doi: #1}\else
	\providecommand{\doi}{doi: \begingroup \urlstyle{rm}\Url}\fi
	
	\bibitem[Mena et~al.(2023)Mena, Coussement, De~Bock, De~Caigny, and
	Lessmann]{mena2023exploiting}
	Gary Mena, Kristof Coussement, Koen~W De~Bock, Arno De~Caigny, and Stefan
	Lessmann.
	\newblock Exploiting time-varying rfm measures for customer churn prediction
	with deep neural networks.
	\newblock \emph{Annals of Operations Research}, pages 1--23, 2023.
	
	\bibitem[Madhav et~al.(2023)Madhav, Ambekar, and Hudnurkar]{madhav2023weld}
	Manu Madhav, Suhas~Suresh Ambekar, and Manoj Hudnurkar.
	\newblock Weld defect detection with convolutional neural network: an
	application of deep learning.
	\newblock \emph{Annals of Operations Research}, pages 1--24, 2023.
	
	\bibitem[Qui{\~n}onero-Candela et~al.(2022)Qui{\~n}onero-Candela, Sugiyama,
	Schwaighofer, and Lawrence]{quinonero2022dataset}
	Joaquin Qui{\~n}onero-Candela, Masashi Sugiyama, Anton Schwaighofer, and Neil~D
	Lawrence.
	\newblock \emph{Dataset shift in machine learning}.
	\newblock Mit Press, 2022.
	
	\bibitem[Guo et~al.(2017)Guo, Pleiss, Sun, and Weinberger]{guo2017calibration}
	Chuan Guo, Geoff Pleiss, Yu~Sun, and Kilian~Q Weinberger.
	\newblock On calibration of modern neural networks.
	\newblock In \emph{International Conference on Machine Learning}, pages
	1321--1330. PMLR, 2017.
	
	\bibitem[Thuy and Benoit(2023)]{thuy2023explainability}
	Arthur Thuy and Dries~F Benoit.
	\newblock Explainability through uncertainty: Trustworthy decision-making with
	neural networks.
	\newblock \emph{European Journal of Operational Research}, 2023.
	\newblock \doi{10.1016/j.ejor.2023.09.009}.
	
	\bibitem[Ovadia et~al.(2019)Ovadia, Fertig, Ren, Nado, Sculley, Nowozin,
	Dillon, Lakshminarayanan, and Snoek]{ovadia2019can}
	Yaniv Ovadia, Emily Fertig, Jie Ren, Zachary Nado, David Sculley, Sebastian
	Nowozin, Joshua Dillon, Balaji Lakshminarayanan, and Jasper Snoek.
	\newblock Can you trust your model's uncertainty? evaluating predictive
	uncertainty under dataset shift.
	\newblock \emph{Advances in neural information processing systems}, 32, 2019.
	
	\bibitem[Huang et~al.(2017)Huang, Li, Pleiss, Liu, Hopcroft, and
	Weinberger]{huang2017snapshot}
	Gao Huang, Yixuan Li, Geoff Pleiss, Zhuang Liu, John~E Hopcroft, and Kilian~Q
	Weinberger.
	\newblock Snapshot ensembles: Train 1, get m for free.
	\newblock \emph{arXiv preprint arXiv:1704.00109}, 2017.
	\newblock \doi{10.48550/arXiv.1704.00109}.
	
	\bibitem[Wen et~al.(2020)Wen, Tran, and Ba]{wen2020batchensemble}
	Yeming Wen, Dustin Tran, and Jimmy Ba.
	\newblock Batchensemble: an alternative approach to efficient ensemble and
	lifelong learning.
	\newblock \emph{arXiv preprint arXiv:2002.06715}, 2020.
	\newblock \doi{10.48550/arXiv.2002.06715}.
	
	\bibitem[Havasi et~al.(2020)Havasi, Jenatton, Fort, Liu, Snoek,
	Lakshminarayanan, Dai, and Tran]{havasi2020training}
	Marton Havasi, Rodolphe Jenatton, Stanislav Fort, Jeremiah~Zhe Liu, Jasper
	Snoek, Balaji Lakshminarayanan, Andrew~M Dai, and Dustin Tran.
	\newblock Training independent subnetworks for robust prediction.
	\newblock \emph{arXiv preprint arXiv:2010.06610}, 2020.
	\newblock \doi{10.48550/arXiv.2010.06610}.
	
	\bibitem[Zhu et~al.(2023)Zhu, Bilal, M{\aa}rtensson, Hanson, Bj{\"o}rkman, and
	Maki]{zhu2023towards}
	Xiaomeng Zhu, Talha Bilal, P{\"a}r M{\aa}rtensson, Lars Hanson, M{\aa}rten
	Bj{\"o}rkman, and Atsuto Maki.
	\newblock Towards sim-to-real industrial parts classification with synthetic
	dataset.
	\newblock In \emph{Proceedings of the IEEE/CVF Conference on Computer Vision
		and Pattern Recognition}, pages 4453--4462, 2023.
	\newblock \doi{10.1109/CVPRW59228.2023.00468}.
	
	\bibitem[Yang et~al.(2021)Yang, Li, Hu, Liang, and Lv]{yang2021deep}
	H~Yang, WD~Li, KX~Hu, YC~Liang, and YQ~Lv.
	\newblock Deep ensemble learning with non-equivalent costs of fault severities
	for rolling bearing diagnostics.
	\newblock \emph{Journal of Manufacturing Systems}, 61:\penalty0 249--264, 2021.
	\newblock \doi{10.1016/j.jmsy.2021.09.009}.
	
	\bibitem[Wu et~al.(2022)Wu, Zhou, Xu, and Lou]{wu2022cs}
	Zengyuan Wu, Caihong Zhou, Fei Xu, and Wengao Lou.
	\newblock A cs-adaboost-bp model for product quality inspection.
	\newblock \emph{Annals of Operations Research}, 308:\penalty0 685--701, 2022.
	\newblock \doi{10.1007/s10479-020-03798-z}.
	
	\bibitem[Abedin et~al.(2021)Abedin, Moon, Hassan, and Hajek]{abedin2021deep}
	Mohammad~Zoynul Abedin, Mahmudul~Hasan Moon, M~Kabir Hassan, and Petr Hajek.
	\newblock Deep learning-based exchange rate prediction during the covid-19
	pandemic.
	\newblock \emph{Annals of Operations Research}, pages 1--52, 2021.
	\newblock \doi{10.1007/s10479-021-04420-6}.
	
	\bibitem[Baradaran~Rezaei et~al.(2023)Baradaran~Rezaei, Amjadian, Sebt, Askari,
	and Gharaei]{baradaran2023ensemble}
	Hirad Baradaran~Rezaei, Alireza Amjadian, Mohammad~Vahid Sebt, Reza Askari, and
	Abolfazl Gharaei.
	\newblock An ensemble method of the machine learning to prognosticate the
	gastric cancer.
	\newblock \emph{Annals of Operations Research}, 328\penalty0 (1):\penalty0
	151--192, 2023.
	\newblock \doi{10.1007/s10479-022-04964-1}.
	
	\bibitem[Du~Jardin(2021)]{du2021forecasting}
	Philippe Du~Jardin.
	\newblock Forecasting bankruptcy using biclustering and neural network-based
	ensembles.
	\newblock \emph{Annals of Operations Research}, 299\penalty0 (1-2):\penalty0
	531--566, 2021.
	\newblock \doi{10.1007/s10479-019-03283-2}.
	
	\bibitem[Cui et~al.(2022)Cui, Wang, Yin, Fan, Dhamotharan, and
	Kumar]{cui2022carbon}
	Shaoze Cui, Dujuan Wang, Yunqiang Yin, Xin Fan, Lalitha Dhamotharan, and Ajay
	Kumar.
	\newblock Carbon trading price prediction based on a two-stage heterogeneous
	ensemble method.
	\newblock \emph{Annals of Operations Research}, pages 1--25, 2022.
	\newblock \doi{10.1007/s10479-022-04821-1}.
	
	\bibitem[Zhang et~al.(2023)Zhang, Li, Zhao, and Wang]{zhang2023carbon}
	Xingmin Zhang, Zhiyong Li, Yiming Zhao, and Lan Wang.
	\newblock Carbon trading and covid-19: a hybrid machine learning approach for
	international carbon price forecasting.
	\newblock \emph{Annals of Operations Research}, pages 1--29, 2023.
	\newblock \doi{10.1007/s10479-023-05327-0}.
	
	\bibitem[Jiang et~al.(2022)Jiang, Jia, Chen, and Chen]{jiang2022two}
	Manrui Jiang, Lifen Jia, Zhensong Chen, and Wei Chen.
	\newblock The two-stage machine learning ensemble models for stock price
	prediction by combining mode decomposition, extreme learning machine and
	improved harmony search algorithm.
	\newblock \emph{Annals of Operations Research}, pages 1--33, 2022.
	\newblock \doi{10.1007/s10479-020-03690-w}.
	
	\bibitem[Zhang et~al.(2022)Zhang, Fleyeh, and Bales]{zhang2022hybrid}
	Fan Zhang, Hasan Fleyeh, and Chris Bales.
	\newblock A hybrid model based on bidirectional long short-term memory neural
	network and catboost for short-term electricity spot price forecasting.
	\newblock \emph{Journal of the Operational Research Society}, 73\penalty0
	(2):\penalty0 301--325, 2022.
	\newblock \doi{10.1080/01605682.2020.1843976}.
	
	\bibitem[Li and Chen(2021)]{li2021entropy}
	Yiheng Li and Weidong Chen.
	\newblock Entropy method of constructing a combined model for improving loan
	default prediction: A case study in china.
	\newblock \emph{Journal of the Operational Research Society}, 72\penalty0
	(5):\penalty0 1099--1109, 2021.
	\newblock \doi{10.1080/01605682.2019.1702905}.
	
	\bibitem[Krauss et~al.(2017)Krauss, Do, and Huck]{krauss2017deep}
	Christopher Krauss, Xuan~Anh Do, and Nicolas Huck.
	\newblock Deep neural networks, gradient-boosted trees, random forests:
	Statistical arbitrage on the s\&p 500.
	\newblock \emph{European Journal of Operational Research}, 259\penalty0
	(2):\penalty0 689--702, 2017.
	\newblock \doi{10.1016/j.ejor.2016.10.031}.
	
	\bibitem[Easaw et~al.(2023)Easaw, Fang, and Heravi]{easaw2023using}
	Joshy Easaw, Yongmei Fang, and Saeed Heravi.
	\newblock Using polls to forecast popular vote share for us presidential
	elections 2016 and 2020: An optimal forecast combination based on ensemble
	empirical model.
	\newblock \emph{Journal of the Operational Research Society}, 74\penalty0
	(3):\penalty0 905--911, 2023.
	\newblock \doi{10.1080/01605682.2022.2101951}.
	
	\bibitem[Gupta et~al.(2023)Gupta, Anand, Gupta, and Koundal]{gupta2023deep}
	Rupesh Gupta, Vatsala Anand, Sheifali Gupta, and Deepika Koundal.
	\newblock Deep learning model for defect analysis in industry using casting
	images.
	\newblock \emph{Expert Systems with Applications}, page 120758, 2023.
	\newblock \doi{10.1016/j.eswa.2023.120758}.
	
	\bibitem[Bilal and Oyedele(2020)]{bilal2020big}
	Muhammad Bilal and Lukumon~O Oyedele.
	\newblock Big data with deep learning for benchmarking profitability
	performance in project tendering.
	\newblock \emph{Expert Systems with Applications}, 147:\penalty0 113194, 2020.
	\newblock \doi{10.1016/j.eswa.2020.113194}.
	
	\bibitem[Zicari et~al.(2022)Zicari, Folino, Guarascio, and
	Pontieri]{zicari2022combining}
	P~Zicari, G~Folino, M~Guarascio, and L~Pontieri.
	\newblock Combining deep ensemble learning and explanation for intelligent
	ticket management.
	\newblock \emph{Expert Systems with Applications}, 206:\penalty0 117815, 2022.
	\newblock \doi{10.1016/j.eswa.2022.117815}.
	
	\bibitem[Poloni et~al.(2022)Poloni, Ferrari, and Initiative]{poloni2022deep}
	Katia~Maria Poloni, Ricardo~Jos{\'e} Ferrari, and Alzheimer’s
	Disease~Neuroimaging Initiative.
	\newblock A deep ensemble hippocampal cnn model for brain age estimation
	applied to alzheimer’s diagnosis.
	\newblock \emph{Expert Systems with Applications}, 195:\penalty0 116622, 2022.
	\newblock \doi{10.1016/j.eswa.2022.116622}.
	
	\bibitem[Pitakaso et~al.(2023)Pitakaso, Khonjun, Nanthasamroeng, Boonmee,
	Kaewta, Enkvetchakul, Gonwirat, Chokanat, Jirasirilerd, and
	Srichok]{pitakaso2023gamification}
	Rapeepan Pitakaso, Surajet Khonjun, Natthapong Nanthasamroeng, Chawis Boonmee,
	Chutchai Kaewta, Prem Enkvetchakul, Sarayut Gonwirat, Peerawat Chokanat,
	Ganokgarn Jirasirilerd, and Thanatkij Srichok.
	\newblock Gamification design using tourist-generated pictures to enhance
	visitor engagement at intercity tourist sites.
	\newblock \emph{Annals of Operations Research}, pages 1--33, 2023.
	\newblock \doi{10.1007/s10479-023-05590-1}.
	
	\bibitem[Han and Li(2022)]{han2022out}
	Te~Han and Yan-Fu Li.
	\newblock Out-of-distribution detection-assisted trustworthy machinery fault
	diagnosis approach with uncertainty-aware deep ensembles.
	\newblock \emph{Reliability Engineering \& System Safety}, 226:\penalty0
	108648, 2022.
	\newblock \doi{10.1016/j.ress.2022.108648}.
	
	\bibitem[Kim et~al.(2023)Kim, Koo, and Hwang]{kim2023unified}
	Jihyo Kim, Jiin Koo, and Sangheum Hwang.
	\newblock A unified benchmark for the unknown detection capability of deep
	neural networks.
	\newblock \emph{Expert Systems with Applications}, 229:\penalty0 120461, 2023.
	\newblock \doi{10.1016/j.eswa.2023.120461}.
	
	\bibitem[Prasad et~al.(2024)Prasad, Deo, Downs, Casillas-P{\'e}rez,
	Salcedo-Sanz, and Parisi]{prasad2024very}
	Salvin~Sanjesh Prasad, Ravinesh~Chand Deo, Nathan~James Downs, David
	Casillas-P{\'e}rez, Sancho Salcedo-Sanz, and Alfio~Venerando Parisi.
	\newblock Very short-term solar ultraviolet-a radiation forecasting system with
	cloud cover images and a bayesian optimized interpretable artificial
	intelligence model.
	\newblock \emph{Expert Systems with Applications}, 236:\penalty0 121273, 2024.
	\newblock \doi{10.1016/j.eswa.2023.121273}.
	
	\bibitem[Wen et~al.(2022)Wen, Xie, Li, and Gao]{wen2022new}
	Long Wen, Xiaotong Xie, Xinyu Li, and Liang Gao.
	\newblock A new ensemble convolutional neural network with diversity
	regularization for fault diagnosis.
	\newblock \emph{Journal of Manufacturing Systems}, 62:\penalty0 964--971, 2022.
	\newblock \doi{10.1016/j.jmsy.2020.12.002}.
	
	\bibitem[Mena et~al.(2021)Mena, Pujol, and Vitri{\`a}]{mena2021survey}
	Jos{\'e} Mena, Oriol Pujol, and Jordi Vitri{\`a}.
	\newblock A survey on uncertainty estimation in deep learning classification
	systems from a bayesian perspective.
	\newblock \emph{ACM Computing Surveys (CSUR)}, 54\penalty0 (9):\penalty0 1--35,
	2021.
	\newblock \doi{10.1145/3477140}.
	
	\bibitem[Zou and Chen(2021)]{zou2021resilience}
	Qiling Zou and Suren Chen.
	\newblock Resilience-based recovery scheduling of transportation network in
	mixed traffic environment: A deep-ensemble-assisted active learning approach.
	\newblock \emph{Reliability Engineering \& System Safety}, 215:\penalty0
	107800, 2021.
	\newblock \doi{10.1016/j.ress.2021.107800}.
	
	\bibitem[Hendrickx et~al.(2024)Hendrickx, Perini, Van~der Plas, Meert, and
	Davis]{hendrickx2024machine}
	Kilian Hendrickx, Lorenzo Perini, Dries Van~der Plas, Wannes Meert, and Jesse
	Davis.
	\newblock Machine learning with a reject option: A survey.
	\newblock \emph{Machine Learning}, 113\penalty0 (5):\penalty0 3073--3110, 2024.
	
	\bibitem[Homenda et~al.(2014)Homenda, Luckner, and
	Pedrycz]{homenda2014classification}
	Wladyslaw Homenda, Marcin Luckner, and Witold Pedrycz.
	\newblock Classification with rejection based on various svm techniques.
	\newblock In \emph{2014 International Joint Conference on Neural Networks
		(IJCNN)}, pages 3480--3487. IEEE, 2014.
	\newblock \doi{10.1109/IJCNN.2014.6889655}.
	
	\bibitem[Kraus et~al.(2020)Kraus, Feuerriegel, and Oztekin]{kraus2020deep}
	Mathias Kraus, Stefan Feuerriegel, and Asil Oztekin.
	\newblock Deep learning in business analytics and operations research: Models,
	applications and managerial implications.
	\newblock \emph{European Journal of Operational Research}, 281\penalty0
	(3):\penalty0 628--641, 2020.
	\newblock ISSN 0377-2217.
	\newblock \doi{10.1016/j.ejor.2019.09.018}.
	\newblock Featured Cluster: Business Analytics: Defining the field and
	identifying a research agenda.
	
	\bibitem[Der~Kiureghian and Ditlevsen(2009)]{der2009aleatory}
	Armen Der~Kiureghian and Ove Ditlevsen.
	\newblock Aleatory or epistemic? does it matter?
	\newblock \emph{Structural safety}, 31\penalty0 (2):\penalty0 105--112, 2009.
	\newblock \doi{10.1016/j.strusafe.2008.06.020}.
	
	\bibitem[Lakshminarayanan et~al.(2017)Lakshminarayanan, Pritzel, and
	Blundell]{lakshminarayanan2017simple}
	Balaji Lakshminarayanan, Alexander Pritzel, and Charles Blundell.
	\newblock Simple and scalable predictive uncertainty estimation using deep
	ensembles.
	\newblock \emph{Advances in neural information processing systems}, 30, 2017.
	
	\bibitem[Thuy et~al.(2024)Thuy, Loginova, and Benoit]{thuy2024active}
	Arthur Thuy, Ekaterina Loginova, and Dries~F Benoit.
	\newblock Active learning to guide labeling efforts for question difficulty
	estimation.
	\newblock \emph{arXiv preprint arXiv:2409.09258}, 2024.
	\newblock \doi{10.48550/arXiv.2409.09258}.
	
	\bibitem[H{\"u}llermeier and Waegeman(2021)]{hullermeier2021aleatoric}
	Eyke H{\"u}llermeier and Willem Waegeman.
	\newblock Aleatoric and epistemic uncertainty in machine learning: An
	introduction to concepts and methods.
	\newblock \emph{Machine Learning}, 110\penalty0 (3):\penalty0 457--506, 2021.
	\newblock \doi{10.1007/s10994-021-05946-3}.
	
	\bibitem[Depeweg et~al.(2018)Depeweg, Hernandez-Lobato, Doshi-Velez, and
	Udluft]{depeweg2018decomposition}
	Stefan Depeweg, Jose-Miguel Hernandez-Lobato, Finale Doshi-Velez, and Steffen
	Udluft.
	\newblock Decomposition of uncertainty in bayesian deep learning for efficient
	and risk-sensitive learning.
	\newblock In \emph{International Conference on Machine Learning}, pages
	1184--1193. PMLR, 2018.
	
	\bibitem[Zhu and Gupta(2017)]{zhu2017prune}
	Michael Zhu and Suyog Gupta.
	\newblock To prune, or not to prune: exploring the efficacy of pruning for
	model compression.
	\newblock \emph{arXiv preprint arXiv:1710.01878}, 2017.
	\newblock \doi{10.48550/arXiv.1710.01878}.
	
	\bibitem[Ernst and Cohen(1990)]{ernst1990operations}
	Ricardo Ernst and Morris~A Cohen.
	\newblock Operations related groups (orgs): a clustering procedure for
	production/inventory systems.
	\newblock \emph{Journal of Operations Management}, 9\penalty0 (4):\penalty0
	574--598, 1990.
	\newblock \doi{10.1016/0272-6963(90)90010-B}.
	
	\bibitem[Hu et~al.(2015)Hu, Bai, Zhao, and Cao]{hu2015modeling}
	Qiwei Hu, Yongsheng Bai, Jianmin Zhao, and Wenbin Cao.
	\newblock Modeling spare parts demands forecast under two-dimensional
	preventive maintenance policy.
	\newblock \emph{Mathematical Problems in Engineering}, 2015, 2015.
	\newblock \doi{10.1155/2015/728241}.
	
	\bibitem[Chang(2023)]{chang2023fuzzy}
	Tsang-Chuan Chang.
	\newblock A fuzzy evaluation approach to determine superiority of deep learning
	network system in terms of recognition capability: case study of lung cancer
	imaging.
	\newblock \emph{Annals of Operations Research}, pages 1--21, 2023.
	\newblock \doi{10.1007/s10479-023-05299-1}.
	
	\bibitem[Nado et~al.(2021)Nado, Band, Collier, Djolonga, Dusenberry, Farquhar,
	Feng, Filos, Havasi, and Jenatton]{nado2021uncertainty}
	Zachary Nado, Neil Band, Mark Collier, Josip Djolonga, Michael~W Dusenberry,
	Sebastian Farquhar, Qixuan Feng, Angelos Filos, Marton Havasi, and Rodolphe
	Jenatton.
	\newblock Uncertainty baselines: Benchmarks for uncertainty \& robustness in
	deep learning.
	\newblock \emph{arXiv preprint arXiv:2106.04015}, 2021.
	\newblock \doi{10.48550/arXiv.2106.04015}.
	
	\bibitem[Fort et~al.(2019)Fort, Hu, and Lakshminarayanan]{fort2019deep}
	Stanislav Fort, Huiyi Hu, and Balaji Lakshminarayanan.
	\newblock Deep ensembles: A loss landscape perspective.
	\newblock \emph{arXiv preprint arXiv:1912.02757}, 2019.
	\newblock \doi{10.48550/arXiv.1912.02757}.
	
	\bibitem[Thuy and Benoit(2024)]{Thuy_Reject_2024}
	Arthur Thuy and Dries~F. Benoit.
	\newblock {Reject}, March 2024.
	\newblock URL \url{https://github.com/arthur-thuy/reject}.
	
	\bibitem[Agrrawal et~al.(2010)Agrrawal, Borgman, Clark, and
	Strong]{agrrawal2010using}
	Pankaj Agrrawal, Richard Borgman, John~M Clark, and Robert Strong.
	\newblock Using the price-to-earnings harmonic mean to improve firm valuation
	estimates.
	\newblock \emph{Journal of Financial Education}, pages 98--110, 2010.
	
	\bibitem[Sokolova and Lapalme(2009)]{sokolova2009systematic}
	Marina Sokolova and Guy Lapalme.
	\newblock A systematic analysis of performance measures for classification
	tasks.
	\newblock \emph{Information processing \& management}, 45\penalty0
	(4):\penalty0 427--437, 2009.
	\newblock \doi{10.1016/j.ipm.2009.03.002}.
	
	\bibitem[Tobin et~al.(2017)Tobin, Fong, Ray, Schneider, Zaremba, and
	Abbeel]{tobin2017domain}
	Josh Tobin, Rachel Fong, Alex Ray, Jonas Schneider, Wojciech Zaremba, and
	Pieter Abbeel.
	\newblock Domain randomization for transferring deep neural networks from
	simulation to the real world.
	\newblock In \emph{2017 IEEE/RSJ international conference on intelligent robots
		and systems (IROS)}, pages 23--30. IEEE, 2017.
	\newblock \doi{10.1109/IROS.2017.8202133}.
	
	\bibitem[Akiba et~al.(2019)Akiba, Sano, Yanase, Ohta, and
	Koyama]{akiba2019optuna}
	Takuya Akiba, Shotaro Sano, Toshihiko Yanase, Takeru Ohta, and Masanori Koyama.
	\newblock Optuna: A next-generation hyperparameter optimization framework.
	\newblock In \emph{Proceedings of the 25th ACM SIGKDD international conference
		on knowledge discovery \& data mining}, pages 2623--2631, 2019.
	\newblock \doi{10.1145/3292500.3330701}.
	
\end{thebibliography}

\newpage
\appendix

\begin{center}
	\textbf{\LARGE Appendix}\\
	\vspace{3mm}
	\textbf{\large Fast and reliable uncertainty quantification with neural network ensembles for industrial image classification}
\end{center}

\section{Dataset}
\label{appendix:dataset}

Table \ref{tab:dataset_sizes_detail} displays a detailed overview of the dataset, per use case and object.

\begin{table}[h!]
	\caption{Number of images per use case and per object}\label{tab:dataset_sizes_detail}
	\centering
	\begin{tabular}{lllccc}  
		\toprule%
		\textbf{Set} & \textbf{Use case} & \textbf{Object} & \textbf{Train} & \textbf{Validation} & \textbf{Test} \\
		\midrule
		\multirow{5}{*}{ID} & \multirow{5}{*}{Use case 1} & Airgun & 1020 & 180 & 300 \\
		& & Electricity12 & 1020 & 180 & 300 \\
		& & Hammer & 1020 & 180 & 300 \\
		& & Hook & 1020 & 180 & 300 \\
		& & Plug & 1020 & 180 & 300 \\
		\cmidrule(lr){1-6}
		\multirow{10}{*}{OOD} & \multirow{3}{*}{Use case 2} & Fork1 & 1020 & 180 & 300 \\
		& & Fork2 & 1020 & 180 & 300 \\
		& & Fork3 & 1020 & 180 & 300 \\
		\cmidrule(lr){2-6}
		& \multirow{4}{*}{Use case 3} & CouplingHalf & \NA & \NA & 300 \\
		& & Gear1 & \NA & \NA & 300 \\
		& & Gear2 & \NA & \NA & 300 \\
		& & Pinion & \NA & \NA & 300 \\
		\cmidrule(lr){2-6}
		& \multirow{3}{*}{Use case 4} & Cross & \NA & \NA & 300 \\
		& & Pin1 & \NA & \NA & 300 \\
		& & Pin2 & \NA & \NA & 300 \\
		\bottomrule
	\end{tabular}
\end{table}

\section{Hyperparameter Tuning}
\label{appendix:hyperparams}

All models are trained using the Adam optimizer and the snapshot ensemble uses the custom cyclic learning rate schedule.
The single NN, deep ensemble, and snapshot ensemble are trained for 200 epochs.
Following \citet{nado2021uncertainty}, the batch ensemble and MIMO ensemble are trained for 50\% longer (i.e., 300 epochs) because they take longer to converge.
The training batch size was set to 512. 
For batch ensemble, we initialize the fast weights to be random sign vectors, as the authors of batch ensemble note that this encourages diversity among the members.

Hyperparameter tuning is used to select the initial learning rate and the L2 regularization weight. 
For the batch ensemble and MIMO ensemble, we tune one additional hyperparameter.
Hyperparameters are tuned independently for each ensemble type, except for the deep ensemble as it simply consists of multiple tuned single NNs.
Table \ref{tab:hyperparams} shows the hyperparameter settings.
Hyperparameters are optimized through a Tree-structured Parzen Estimator using Optuna \citep{akiba2019optuna} for 20 trials.
We follow \citet{ovadia2019can} and minimize the NLL on the validation set; they optimize NLL rather than accuracy since the former is a proper scoring rule.
Table \ref{tab:val_nll} shows the optimal validation NLL obtained during hyperparameter tuning.
In the main results section, we work with a snapshot ensemble of $M=8$ members and a MIMO ensemble of $M=3$ members because this yields the lowest validation NLL for their ensemble type.

\begin{table}[h!]
	\caption{\label{tab:hyperparams}Hyperparameter values}
	\centering
	\begin{tabular}{lllc}\toprule 
		& \textbf{Hyperparameter} & \textbf{Range} & \textbf{Selected value} \\
		\midrule
		\multicolumn{4}{l}{\small \textsc{Single}} \\
		& Epochs & \{200\} & 200 \\
		& Initial learning rate & [\num{1e-4}, \num{5e-2}] & \num{0.007586282912758596} \\
		& L2 penalty & [\num{1e-4}, \num{1e-1}] & \num{0.00031344095561317286} \\
		\multicolumn{4}{l}{\small \textsc{Snapshot ensemble ($M = 4$)}} \\
		& Epochs & \{200\} & 200 \\
		& Initial learning rate & [\num{1e-4}, \num{5e-2}] & \num{0.011457405014309335} \\
		& L2 penalty & [\num{1e-5}, \num{1e-1}] & \num{0.00020652854145967596} \\
		\multicolumn{4}{l}{\small \textsc{Snapshot ensemble ($M = 6$)}} \\
		& Epochs & \{200\} & 200 \\
		& Initial learning rate & [\num{1e-4}, \num{5e-2}] & \num{0.0029668598676528242} \\
		& L2 penalty & [\num{1e-5}, \num{1e-1}] & \num{0.028438406320732502} \\
		\multicolumn{4}{l}{\small \textsc{Snapshot ensemble ($M = 8$)}} \\
		& Epochs & \{200\} & 200 \\
		& Initial learning rate & [\num{1e-4}, \num{5e-2}] & \num{0.003375136675278424} \\
		& L2 penalty & [\num{1e-5}, \num{1e-1}] & \num{0.00010579456188450328} \\
		\multicolumn{4}{l}{\small \textsc{Batch ensemble ($M = 4$)}} \\
		& Epochs & \{300\} & 300 \\
		& Initial learning rate & [\num{1e-4}, \num{5e-2}] & \num{0.0011109584915055496} \\
		& L2 penalty & [\num{1e-5}, \num{1e-2}] & \num{0.00013166781723713105} \\
		& Fast weight initialisation & \{``random sign''\} & ``random sign'' \\
		& Fast weight LR multiplier & [0.1, 1] & \num[exponent-mode=input]{0.4900455429086983} \\
		\multicolumn{4}{l}{\small \textsc{Batch ensemble ($M = 8$)}} \\
		& Epochs & \{300\} & 300 \\
		& Initial learning rate & [\num{1e-4}, \num{5e-2}] & \num{0.002051399683699425} \\
		& L2 penalty & [\num{1e-5}, \num{1e-2}] & \num{0.00016721222946668377} \\
		& Fast weight initialisation & \{``random sign''\} & ``random sign'' \\
		& Fast weight LR multiplier & [0.1, 1] & \num[exponent-mode=input]{0.3896106424999362} \\
		\multicolumn{4}{l}{\small \textsc{MIMO ensemble ($M = 3$)}} \\
		& Epochs & \{300\} & 300 \\
		& Initial learning rate & [\num{1e-4}, \num{1e-2}] & \num{0.007615313992857735} \\
		& L2 penalty & [\num{1e-7}, \num{1e-2}] & \num{0.00018068343404486047} \\
		& Input repetition probability & [0.0, 0.8] & 0.3 \\
		\multicolumn{4}{l}{\small \textsc{MIMO ensemble ($M = 4$)}} \\
		& Epochs & \{300\} & 300 \\
		& Initial learning rate & [\num{1e-4}, \num{1e-2}] & \num{0.0010379115268074602} \\
		& L2 penalty & [\num{1e-7}, \num{1e-2}] & \num{0.0000014291715250459657} \\
		& Input repetition probability & [0.0, 0.8] & 0.8 \\
		\bottomrule
	\end{tabular}
\end{table}

\begin{table}[h!]
	\caption{\label{tab:val_nll}Optimal validation NLL}
	\centering
	\begin{tabular}{llc}\toprule 
		\textbf{Model} & \textbf{Ensemble size $M$} & \textbf{Val NLL} \\
		\midrule
		Single & \NA & 0.1040\\
		\cmidrule(lr){1-3}
		\multirow{3}{*}{Snapshot ensemble} & 4 & 0.1119\\
		& 6 & 0.1178\\
		& 8 & \textbf{0.1042}\\
		\cmidrule(lr){1-3}
		\multirow{2}{*}{Batch ensemble} & 4 & 0.0460\\
		& 8 & 0.0494\\
		\cmidrule(lr){1-3}
		\multirow{2}{*}{MIMO ensemble} & 3 & \textbf{0.1023}\\
		& 4 & 0.1255\\
		\bottomrule
	\end{tabular}
\end{table}

\section{Results for all ensemble configurations}
\label{appendix:all_results}

This section presents results for all ensemble configurations.
The ensemble results reported in the main body are replicated to provide easier comparisons.

Table \ref{tab:id_predictive_all} shows the classification accuracy and the NLL metrics on the ID set.
For deep and batch ensemble, the difference between ensemble sizes 4 or 8 is minimal.
For snapshot and MIMO ensemble, the configuration selected based on the validation NLL ($M=8$ for snapshot and $M=3$ for MIMO) is on par or better than the other ensemble sizes.

\begin{table}[h!]
	\caption{Performance on the ID set (all configurations)}\label{tab:id_predictive_all}
	\centering
	\begin{tabular}{llcc}\toprule
		\textbf{Model} & \textbf{Ensemble size $M$} & \textbf{Accuracy (\%)} $\uparrow$ & \textbf{NLL} $\downarrow$ \\
		\midrule
		Single & \NA & 97.72 $\pm$ 0.14 & 0.1047 $\pm$ 0.0089 \\
		\cmidrule(lr){1-4}
		\multirow{2}{*}{Deep ensemble} & 4 & 98.51 $\pm$ 0.09 & 0.0496 $\pm$ 0.0027\\
		& 8 & 98.53 $\pm$ 0.09 & 0.0440 $\pm$ 0.0011 \\
		\cmidrule(lr){1-4}
		\multirow{3}{*}{Snapshot ensemble} & 4 & 96.68 $\pm$ 0.26 & 0.1029 $\pm$ 0.0065 \\
		& 6 & 95.20 $\pm$ 0.16 & 0.1526 $\pm$ 0.0030\\
		& 8 & 97.15 $\pm$ 0.10 & 0.1048 $\pm$ 0.0031 \\
		\cmidrule(lr){1-4}
		\multirow{2}{*}{Batch ensemble} & 4 & 98.47 $\pm$ 0.07 & 0.0514 $\pm$ 0.0026 \\
		& 8 & 98.65 $\pm$ 0.05 & 0.0504 $\pm$ 0.0035 \\
		\cmidrule(lr){1-4}
		\multirow{2}{*}{MIMO ensemble} & 3 & 96.63 $\pm$ 0.30 & 0.1057 $\pm$ 0.0061 \\
		& 4 & 96.57 $\pm$ 0.16 & 0.1085 $\pm$ 0.0020\\
		\bottomrule
	\end{tabular}
\end{table}

\begin{figure}[h!]
	\centering
	\includegraphics[width=0.6\textwidth]{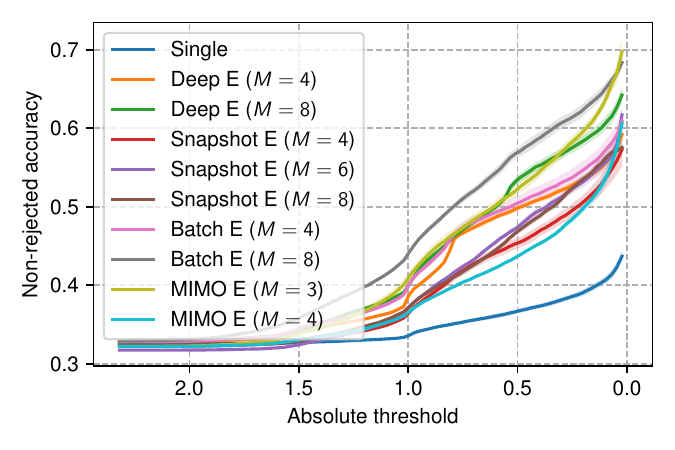}
	\caption{Non-rejected accuracy (all configurations)}
	\label{fig:nra_all}
\end{figure}

Figure \ref{fig:nra_all} displays the NRA for all ensemble configurations based on the total uncertainty.
For deep and batch ensemble, we observe a sizable decrease in NRA when transitioning from 8 ensemble members to only 4.
Specifically, at lower uncertainty thresholds, the NRA curves with $M=4$ are 5--8 percentage points lower than those with $M=8$.
This finding shows that although configurations with $M=4$ show similar ID performance than with $M=8$, the larger number of members yields better performance on the OOD set.
For snapshot ensemble, the selected configuration $M=8$ performs on par or better than the other ensemble sizes.
The selected MIMO configuration $M=3$ outperforms the other ensemble size $M=4$ by a significant margin.

\begin{figure}[h!]
	\centering
	\includegraphics[width=0.6\textwidth]{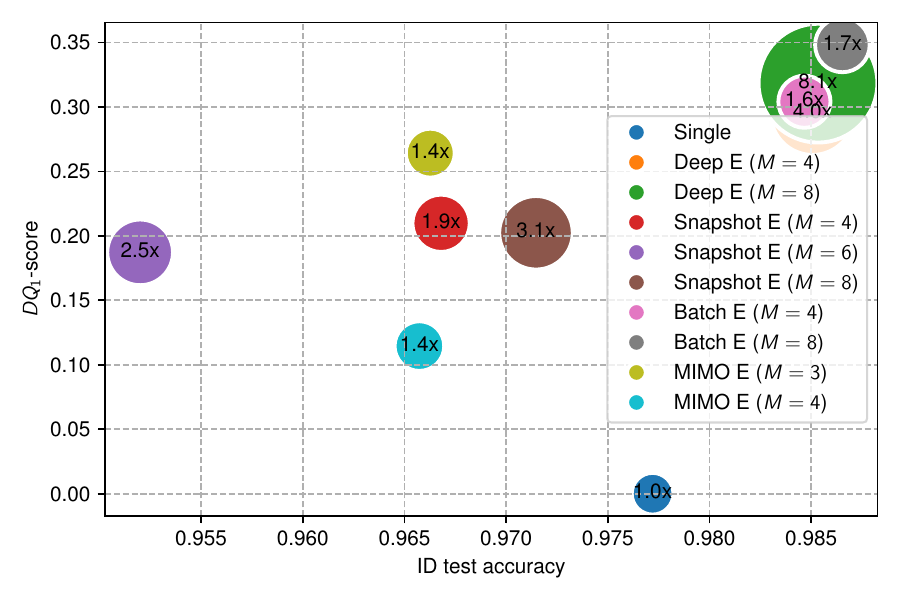} 
	\caption{Accuracy on the ID set and $DQ_1$-score (all configurations).
		Each point represents an ensemble model; the point size represents the weighted computational cost relative to a single NN.
		Models in the upper-right corner perform best}
	\label{fig:cost_performance_all}
\end{figure}

Figure \ref{fig:cost_performance_all} displays a bubble chart combining the predictive accuracy, diversity scores, and computational costs of ensemble configurations.
Similarly to the $M=8$ configuration, for $M=4$ batch ensemble exhibits comparable performance to the deep ensemble, despite being much less computationally expensive.
For snapshot ensemble, the alternative configurations with $M=4$ and $M=6$ are inferior to the selected configuration $M=8$ as the $DQ_1$-scores are equal but the accuracy values are lower.
The alternative MIMO configuration with $M=4$ is inferior to the selected $M=3$ configuration as the accuracy is equal but the $DQ_1$-score is much worse.

As such, we conclude that deep and batch ensemble significantly outperform the other models, showcasing much higher ID accuracy and $DQ_1$-scores.
Batch ensemble is substantially more efficient than deep ensemble and forms an excellent alternative.

\section{$DQ_\beta$-score sensitivity}
\label{appendix:beta_sensitivity}

\begin{figure}[h!]
	\centering
	\begin{subfigure}[b]{0.60\textwidth}
		\includegraphics[width=1\linewidth]{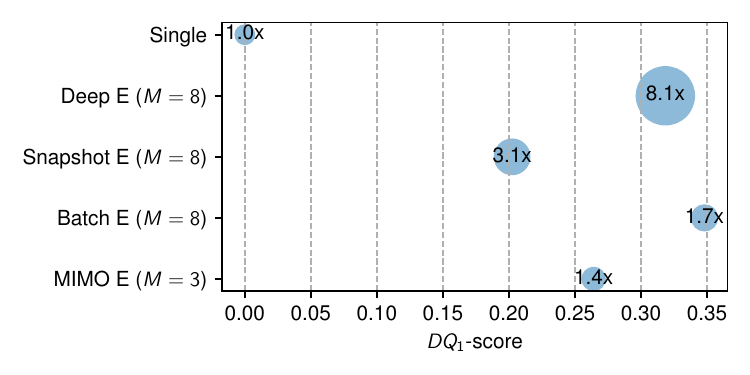}
		\caption{$DQ_1$-score}
		\label{fig:dq_beta1} 
	\end{subfigure}
	
	\begin{subfigure}[b]{0.60\textwidth}
		\includegraphics[width=1\linewidth]{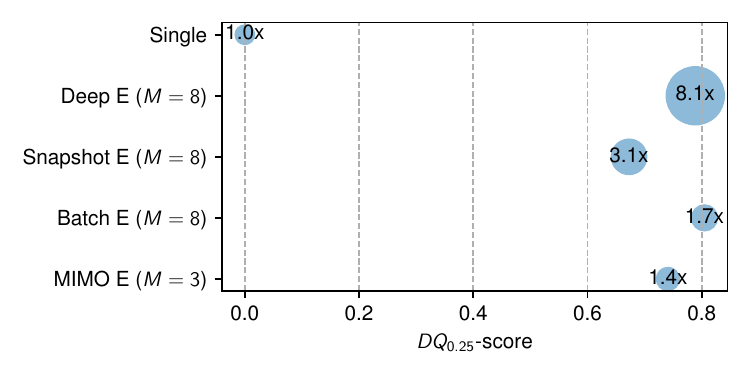}
		\caption{$DQ_{0.25}$-score}
		\label{fig:dq_beta0.25}
	\end{subfigure}
	
	\begin{subfigure}[b]{0.60\textwidth}
		\includegraphics[width=1\linewidth]{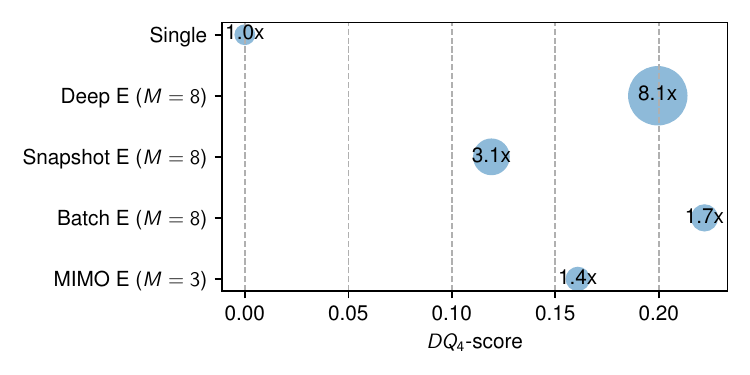}
		\caption{$DQ_4$-score}
		\label{fig:dq_beta4}
	\end{subfigure}
	
	\caption{$DQ_\beta$-scores for varying $\beta$ values}
	\label{fig:dq_beta}
\end{figure}

Figure \ref{fig:dq_beta} displays a bubble chart of the $DQ_\beta$-scores for varying $\beta$ values, where the bubble size denotes the computational expense as discussed in Subsection \ref{subsec:cost_performance}.
Subfigure \ref{fig:dq_beta1} shows the $DQ_1$-score, which corresponds to the vertical axis of Figure \ref{fig:cost_performance}.

Subfigure \ref{fig:dq_beta0.25} shows the results for $\beta=0.25$, indicating that the ID performance is 4 times more important to the practitioner than the OOD performance.
The bubbles are now closer to each other because the largest performance differences are observed on the OOD set, which is now weighted less in the $DQ$-score.
Note that the $DQ$-scores are now substantially higher as the IDD scores are higher than the OODD scores.

Subfigure \ref{fig:dq_beta4} shows the results for the parameter $\beta=4$, reflecting the practitioner's priority for OOD performance over ID performance.
In contrast to before, the bubbles now move further apart because we observe larger differences in OODD.
For example, the $DQ$-score of the snapshot ensemble is now lower relative to the deep ensemble as the snapshot ensemble performed poorly in terms of OODD.

The ranking of the methods remains unchanged, but the relative differences between them do shift. 
The $DQ$-score helps practitioners more easily assess whether the potential improvement in diversity performance justifies the development of a more computationally intensive model. 
Management can set the $\beta$ value, while the machine learning team can then choose the most suitable model type for production.

\end{document}